\documentclass[11pt]{article}
\usepackage[preprint]{acl}
\usepackage{times}
\usepackage{latexsym}
\usepackage[T1]{fontenc}
\usepackage[utf8]{inputenc}
\usepackage{microtype}
\usepackage{courier}
\usepackage{graphicx}
\usepackage{placeins}
\usepackage{booktabs,tabularx}
\usepackage{amsmath,amssymb}
\hypersetup{pdftitle={Deeper Backpropagation or More Updates? Training Looped Language Models Across Budgets},pdfauthor={Jiarun Shen, Xiao Li}}
\title{Deeper Backpropagation or More Updates?\\Training Looped Language Models Across Budgets}
\author{Jiarun Shen \quad Xiao Li \\
  \normalfont The Chinese University of Hong Kong, Shenzhen \\
  \normalfont\texttt{jiarunshen@link.cuhk.edu.cn} \quad \texttt{lixiao@cuhk.edu.cn}}
\newcommand{\K}[1]{\ensuremath{K#1}}
\newcommand{\Te}{\ensuremath{T_{\mathrm{eval}}}}

\newcounter{algorithm}
\newsavebox{\papertablebox}
\newcommand{\papertable}[1]{\begingroup\small\setlength{\tabcolsep}{4pt}%
\sbox{\papertablebox}{#1}%
\ifdim\wd\papertablebox>\linewidth
\resizebox{\linewidth}{!}{\usebox{\papertablebox}}%
\else\usebox{\papertablebox}\fi\endgroup}
\begin{document}
\maketitle
\raggedbottom
\begin{abstract}
Repeating transformer layers can improve language model performance without adding parameters.
However, the added computational depth increases both forward and backward computation costs during training.
Training such models requires choosing a backward window that controls how many final layer repetitions retain gradients during backpropagation.
Longer windows improve per-token learning efficiency but increase per-update cost, while shorter windows reduce per-update cost but sacrifice per-token efficiency.
Understanding which window performs better requires examining different training budgets.
Prior work compares backward windows across computational budgets.
When training budgets increase, it remains unclear whether the same window continues to perform best.
We measure how the relative performance of two windows shifts as training budgets grow.
Our method isolates per-token efficiency by training both windows on identical data, and compares complete training recipes at approximately equal training time at each budget.
Results show that shorter windows lose much of their advantage as budgets grow.
Controlled experiments show that both benefits decrease at larger budgets, with a greater decline in the benefit of extended training.
Backward window comparisons depend on the available budget, and the preferred window among the tested candidates can change as the budget grows.
\end{abstract}

\section{Introduction}
Repeating transformer layers can improve language model performance without adding parameters \citep{dehghani2019universal,prairie2026parcae}. Each pass through the shared layers constitutes one recurrence iteration, producing multiple stages of computation from the same parameters. Repeating layers increases computational depth, which raises both forward and backward computation costs during training. Training such models requires choosing how many recurrence iterations to retain for backpropagation. We define the backward window as the maximum number of trailing recurrence iterations through which gradients propagate. Truncating the backward window reduces backward computation cost while preserving all forward passes. Architecture and inference computation remain unchanged regardless of which backward window is used during training.

Choosing a backward window determines how training time divides between computation per update and total number of updates. Deeper backpropagation improves per-token efficiency when training data is held fixed. Shorter windows reduce per-update cost, permitting more updates within a fixed training-time budget. Two factors govern backward window selection. First, per-token efficiency measures how much deeper backpropagation improves learning from a fixed quantity of training data. Second, gains from extended training quantify the benefit of running additional optimization updates. When training budgets increase, per-token efficiency and extended training gains may shift by different magnitudes. Understanding backward window selection requires examining whether a window that performs best under limited compute continues to perform best when more compute becomes available.

Prior work establishes components of the backward window trade-off. Parcae compares backward windows at a fixed token count and documents per-token efficiency gains from deeper backpropagation \citep{prairie2026parcae}. \citet{schwethelm2026recurrence} examine truncated backpropagation across multiple compute budgets, including how saved computation supports additional training tokens. We build on this work by measuring matched-time differences at multiple budgets with paired runs that share random seeds, and decomposing these differences using additional equal-token controls. Paired comparisons at each budget measure how shorter windows perform when allocated more updates within the same training time. Equal-token controls isolate per-token efficiency from extended training contributions, quantifying how each factor shifts with budget scale.

We conduct two complementary comparisons at each budget level. Equal-token comparisons train models with both backward windows on identical training data and measure per-token efficiency differences. Matched-time comparisons allocate approximately equal training time to each backward window, giving the shorter window more updates within the same time budget. The first comparison isolates per-token efficiency by holding data fixed. The second comparison measures combined gains from both per-token efficiency and extended training. Extended training here refers to a complete training recipe with additional optimization updates and a schedule adjusted for the larger update count.

We apply this method to backward windows of four and eight recurrence iterations in a looped language model. Model architecture, forward recurrence iteration distribution during training, and inference iteration count remain fixed within each paired comparison. Each comparison uses the same random seed for both backward windows. The main experiment compares a baseline training budget with a budget four times larger. Measured per-update costs determine how many additional updates the shorter window receives in matched-time comparisons.

At the smaller budget, the shorter window trained with additional updates achieves lower test loss than the longer window across all training seeds. At the larger budget, this advantage shrinks in all seeds evaluated at both budgets. Controlled experiments show that per-token efficiency from deeper backpropagation and gains from extended training both diminish at larger budgets. Extended training gains diminish more, explaining the shift in matched-time comparisons. Training budget shapes the balance between backward window depth and extended training.

Our contributions are:
\begin{itemize}
\item Empirical evidence that shorter backward windows consistently see diminished matched-time advantage as training budgets grow, demonstrated across training seeds, learning rates, and model configurations.
\item Controlled decomposition of matched-time shifts into per-token efficiency and extended training gains, showing that extended training gains diminish more than per-token efficiency gains at larger budgets.
\end{itemize}

\section{Related Work}
\paragraph{Backward-window ablations.}
Universal Transformers introduced recurrence over shared transformations \citep{dehghani2019universal}. Public recurrent-depth models reuse shared blocks to increase computation without proportionally increasing unique parameters. Huginn truncates backpropagation to the last eight recurrent steps to control training compute and memory \citep{geiping2025latent}. We use Parcae's architecture and optimizer framework \citep{prairie2026parcae}. Parcae's Appendix~I, Figure~16 and Table~10 vary backward windows from 4 to 12 at mean forward recurrence 20 and 8.5B training tokens. Parcae establishes per-token efficiency gains from deeper backpropagation at fixed token counts. We examine whether per-token efficiency gains justify their cost when shorter windows use saved time for additional updates, and whether this trade-off shifts across training budgets.

\paragraph{Compute allocation and recurrence value.}
\citet[\S5.1]{schwethelm2026recurrence} study truncated backpropagation across compute budgets and widths, allocating saved computation to additional tokens. Their study already includes multi-budget truncation experiments and fixed-model comparisons. We measure the same-seed change in the relative loss of two backward windows, test its sensitivity to learning-rate selection and model configuration, and add equal-token endpoints to determine whether the window gap itself grows. These controls clarify the observed contraction beyond the established use of truncation savings for more tokens. The comparisons do not provide a predictive scaling law.

\paragraph{Gradient estimation and evaluation.}
Adaptive truncation controls gradient bias \citep{aicher2020adaptive}; randomized truncation and telescoping construct compensated estimators \citep{tallec2017unbiasing,beatson2019loops}. We compare completed fixed-window recipes. Pretraining loss and downstream performance may not align \citep{liu2023implicit,saunshi2025latent}, so downstream scores provide additional evidence beyond NLL comparisons. We distinguish training-seed variation \citep{sellam2022multiberts} from block-resampling uncertainty that conditions on fixed checkpoints \citep{dror2018testing}.

\section{Experimental Design}
\label{sec:design}
We measure language modeling by negative log-likelihood (NLL) in nats per token, where lower is better. For downstream evaluation, \textbf{CORE} is the mean chance-adjusted accuracy across 22 tasks. CORE$\times100$ has 0 at the tasks' chance baselines and 100 at perfect accuracy. The CORE metric is distinct from the model's shared \emph{Core} layers. LAMBADA measures final-word accuracy. Higher downstream scores are better.

\subsection{Models and backward windows}
The primary Parcae configuration has 144,323,136 parameters and width 768. Two input Transformer blocks (the Prelude) produce representation $e$ from token embeddings. Each iteration applies the same two shared Core blocks to the recurrent state, injecting $e$ at each step. Two output blocks (the Coda) and a prediction head produce next-token logits. The second configuration has 388,003,328 parameters, width 1,024, and four blocks each in the Prelude, Core, and Coda.

\begin{figure*}[t]
\centering\includegraphics[width=\textwidth]{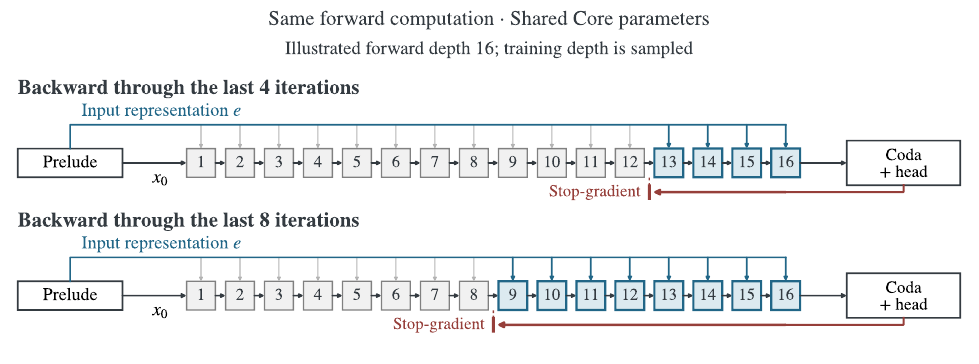}
\caption{\textbf{Four- and eight-iteration backward windows with the same forward computation.} Both rows illustrate 16 forward iterations to the shared two-block Core. The Prelude produces input representation $e$, and $x_0$ denotes the initial recurrent state. Each Core iteration updates the state using an injection of $e$. The Coda and prediction head map the final state to next-token logits. K4 retains the gradient graph for iterations 13--16, and K8 for iterations 9--16. Gray iterations execute in the forward pass without retaining a gradient graph. Red arrows show backpropagation through the retained iterations. Stop-gradient detaches the state before the retained window, while injections into retained iterations preserve gradients to the Prelude. Gradients from retained iterations accumulate into the shared Core parameters. Training samples forward depth $T_i$, so a window $k$ retains $\min(k,T_i)$ iterations. One optimizer update follows gradient accumulation across microbatches.}
\label{fig:graph}
\end{figure*}

Training draws per-sequence forward depths $T_i=\max(1,\mathrm{Poisson}(16))$. A backward window $k$ retains gradients through the final $\min(k,T_i)$ active iterations. Earlier iterations execute in the forward pass without retaining gradients. We detach the recurrent state at the boundary; retained injections of $e$ preserve gradient paths to the Prelude. All original parameters remain trainable. Architecture and sampled forward depths remain fixed within each configuration, and primary evaluation uses \Te$=16$. Four-, six-, and eight-iteration backpropagation are denoted K4, K6, and K8. FULL retains every sampled recurrence. Figure~\ref{fig:graph} illustrates backward window truncation; Algorithm~\ref{alg:window} details batching.

Training uses a document-deduplicated packed FineWeb-Edu subset \citep{penedo2024fineweb}, the Parcae tokenizer, and sequence length 1,024. Each update accumulates 16 microbatches of eight sequences, consuming 131,072 targets. The longest run uses 2,633,891,840 targets from the 5.088B-token stream. Appendix~\ref{app:recipe} records data, licenses, hyperparameters, and software.

\subsection{Training recipes and paired budgets}
We use native Muon/AdamW, BF16 autocast, FP32 parameters, and norm-one clipping. The original common learning-rate scale 0.008 was selected on pilot seed 1101 using K4 and FULL dev loss at 1,024 updates. Learning rate is constant for the first half and then decays linearly; Muon weight decay declines throughout. Each run starts from a seed-specific random initialization with an empty optimizer and cursor zero, using a complete schedule for the run's total update count. Within each seed, different backward windows share data prefixes, forward-depth draws, and selection/RNG streams.

Eight A800 80GB GPUs run independent single-GPU jobs. Controlled update costs $c_k$ measure timing across changing resident batches, two opposite method orders, and 100 updates per timing run, excluding the first 20. Before quality readout we set
\begin{equation}
N_k=\operatorname{round}_{\rm half\mbox{-}up}(N_{\rm ref}c_{\rm ref}/c_k).
\end{equation}
The 144M short-budget comparison trains three models independently. K4 and K8 each train for 4,096 updates in the equal-token comparison. Extended K4 trains for 4,986 updates, using saved backward computation time to add updates while approximately matching K8 training time. Medium and long budgets pair K8 horizons of 8,192 and 16,384 updates with extended K4 horizons of 9,972 and 19,944. We examine whether the short-budget preference persists at larger budgets. We call these budgets $1\times$, $2\times$, and $4\times$, and abbreviate extended four-iteration training as K4+. The 388M short/long K4+ horizons are 5,024/20,095. Each configuration has different per-update costs, so $1\times$ budget corresponds to different absolute training times across configurations. Measured time ratios between K4+ and K8 range from 0.9985 to 1.0074; realized time mismatches across completed comparisons range from $-0.270\%$ to $+1.380\%$. Timing includes the data path and excludes evaluation, and is not FLOP or energy matching.

The common-rate budget comparison uses four paired 144M seeds at $1\times$ and $4\times$, with learning rate 0.008. To assess learning-rate sensitivity, a 24-endpoint grid crosses two windows, two budgets, three rates $\{0.004, 0.008, 0.016\}$, and two exploratory seeds. For each window and budget, mean final dev loss over the two seeds selects the rate before evaluation on two separate confirmation seeds. Two further seeds repeat the budget comparison in the 388M configuration at common rate 0.004, obtained from a 200-update coarse calibration rather than full-horizon independent tuning.

Additional equal-token K4 endpoints complete three-seed controls in the 144M configuration at rate 0.008. Together with K8 and K4+ endpoints, these measure the equal-token benefit of deeper backpropagation and the gain from the complete extended-training recipe. A three-seed K4/K6/K8 comparison examines an intermediate window at approximately matched small-budget time. All primary evaluations use 16 forward iterations. Some analyses reuse trained models and seeds, so they are not additional independent replications. Appendix~\ref{app:software} lists the full seed coverage.

\subsection{Evaluation and uncertainty}
Dev and test NLL use 1,950,720 and 10,485,760 targets. All 36 new endpoints have fixed-T16 dev and test results. These endpoints add neither downstream evaluations nor depth curves to the original coverage (Appendix~\ref{app:eval}). New loss intervals use 20,000 shared paired draws of contiguous 65,536-target blocks, weighting the partial dev block by its target count. Intervals are conditional on checkpoints and do not measure population variation over training seeds or unknown document/book clusters. Earlier test-block and downstream source-cluster analyses retain their original definitions (Appendix~\ref{app:uncertainty}).

CORE averages prompt seeds 1234/2345/3456, reusing identical zero-shot predictions. The label \emph{CORE (adapted)} identifies repaired option boundaries for three prefix-degenerate PIQA items. Independent LAMBADA uses all 5,153 examples, no BOS, and teacher-forced exact final-word correctness \citep{paperno2016lambada}. LAMBADA sources overlap CORE-internal LAMBADA. Prompt seeds are not training replications, and repeated use of the test distribution is not independent benchmark data.

\paragraph{Reproduction scope.}
Paired-state restoration and the training-to-evaluation path were checked in a clean worktree within the existing software environment. Archived paired statistics were also reconstructed. These checks assess the implementation, rather than a clean installation or a full scientific rerun. The accompanying software supplement provides training and evaluation entry points, configurations, statistical routines, and small result tables. Appendix~\ref{app:software} distinguishes those materials from weights, original data, and large numerical inputs that remain outside the package.

\section{Small-Budget Results}
\label{sec:short}
\begin{table*}[t]
\renewcommand{\arraystretch}{1.4}
\centering\papertable{
\begin{tabular}{@{}lrrrrrr@{}}
\toprule
Training recipe & Updates & Targets (M) & Time / K8 & Test NLL & CORE$\times100$ & LAMB. (\%) \\
\midrule
\shortstack[l]{Four-loop backpropagation} & 4,096 & 536.87 & 0.8208 & 3.2937 & 9.404 & 21.813 \\
\shortstack[l]{Eight-loop backpropagation} & 4,096 & 536.87 & 1.0000 & 3.2700 & 10.405 & 23.055 \\
\shortstack[l]{Four-loop backpropagation,\\longer training} & 4,986 & 653.52 & 1.0015 & 3.2523 & 10.150 & 22.679 \\
\bottomrule
\end{tabular}}
\caption{\textbf{Equal tokens favor K8, whereas approximately matched training time favors extended K4 on test loss.} K4+ denotes extended four-iteration training. Each row averages the same three seeds 1102/1103/1104. The first two rows use 4,096 updates; the second and third approximately match time. Targets are millions of training targets; Time/K8 is the mean same-seed time ratio. Test NLL is in nats/token and lower is better. CORE$\times100$ is the adapted chance-adjusted score, not percentage accuracy; LAMBADA is teacher-forced final-word accuracy (\%). Higher task scores are better. All evaluations use 16 inference iterations.}
\label{tab:allocation}
\end{table*}

At 4,096 updates, K8 has lower test NLL than K4 in each exploratory seed, with a mean gap of 0.0237 nats/token (Table~\ref{tab:allocation}). Training K4 uses about 82\% of K8's training time. Using the saved time for additional four-iteration updates changes the NLL ordering. K4+ at 4,986 updates has mean test NLL 3.2523, compared with 3.2700 for K8 at 4,096.

The small-budget result repeats in seeds 1105/1106. K4+ minus K8 test NLL is $-0.0111$ for seed 1105 and $-0.0138$ for seed 1106. Across all five seeds, extending K4 improves small-budget test NLL by 0.011--0.020 nats/token. Extended K4 training in the 144M configuration processes 21.73\% more targets with a longer full optimization schedule. Small-budget efficiency reflects a complete training recipe, rather than additional gradient paths or token exposure alone.

\section{Results Across Budgets}
\label{sec:long}
\subsection{Same-seed budget changes}
\begin{figure*}[t]
\centering\includegraphics[width=\textwidth]{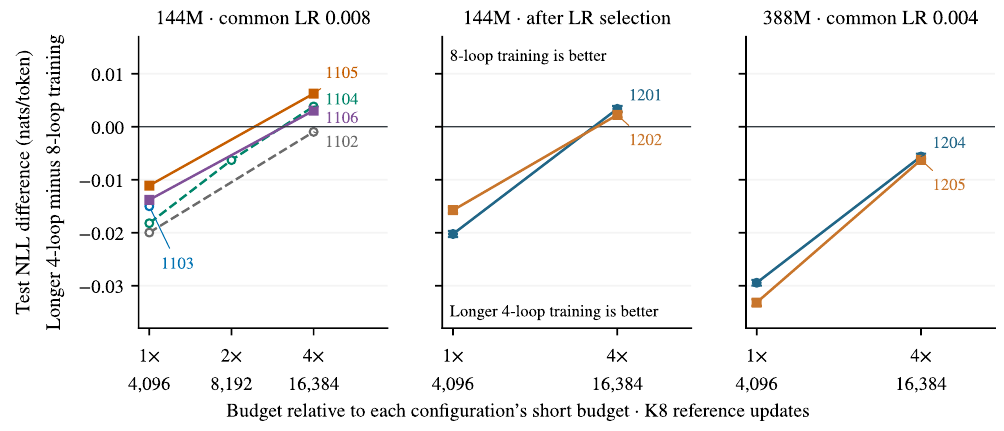}
\caption{\textbf{The matched-time advantage of shorter backpropagation diminishes between the tested budgets, while the better of the two windows at the larger budget varies.} Negative values favor extended four-iteration training (K4+); positive values favor eight-iteration training (K8). The original 144M comparison retains all ten points, including seed 1103 at only the small budget and the medium budget for seed 1104. In the left panel, dashed lines and open markers denote exploratory seeds; solid lines and filled markers denote replication seeds. After dev-based learning-rate selection, both 144M confirmation seeds favor K8 at $4\times$. Both 388M seeds continue to favor K4+ despite diminishing advantage. Lines join separately trained endpoints from the same seed, not checkpoints from one run. The 144M K4+ horizons are 4,986/9,972/19,944; 388M uses 5,024/20,095. Relative budgets do not imply equal absolute GPU time across configurations. All points use T16. Error bars in the middle and right panels are marginal 95\% paired test-block intervals conditional on checkpoints. The original panel preserves its point estimates; intervals are tabulated separately.}
\label{fig:persistence}
\end{figure*}

Figure~\ref{fig:persistence} retains every original matched-time pair. We write $C=L_{\mathrm{K4+}}-L_{\mathrm{K8}}$ and $\Delta C=C_{4\times}-C_{1\times}$. Negative $C$ favors K4+, whereas positive $\Delta C$ denotes contraction of that advantage. The K4+ advantage diminishes in all four seeds evaluated at small and large budgets. Large-budget K4+ minus K8 test NLL is $-0.0010$, $+0.0038$, $+0.0062$, and $+0.0031$ for seeds 1102, 1104, 1105, and 1106. Seed 1102 still favors K4+; the other three favor K8. Seed 1104's medium-budget value is $-0.0063$. The consistent finding across seeds is diminishing advantage, not uniform reversal.

\subsection{Learning-rate and configuration checks}
The complete 144M grid shows positive budget changes in all six seed/rate comparisons, from 0.0185 to 0.0271 test nats/token. The grid includes two exploratory seeds tested at three rates each, yielding six seed/rate pairs that are not independent training replications (Table~\ref{tab:robust-test}). Dev selection chooses 0.016 for both windows at the small budget and 0.008 for both at the large budget. The small-budget choice is at the upper grid boundary.

Table~\ref{tab:budget-confirmation} reports the budget changes for selected-rate confirmation and the second configuration. Both selected 144M comparisons show $\Delta C > 0$ (crossing zero from negative to positive $C$), whereas both 388M comparisons show $\Delta C > 0$ but $C$ remains negative at the large budget. A common direction of contraction therefore does not imply a common large-budget winner. Each 144M window is selected separately; the selected rates happen to coincide. The 388M comparison extends configuration coverage using coarse learning-rate calibration, rather than repeating the full 144M tuning procedure. Seed 1102 at rate 0.008 and seed 1104 at rate 0.004 retain a large-budget K4+ advantage. Contraction is therefore more consistent across the common-rate grid than reversal. Appendix~\ref{app:lr-snapshot} supplies all endpoints, intervals, selections, and realized costs.

\begin{table*}[t]
\renewcommand{\arraystretch}{1.4}
\centering
\small
\setlength{\tabcolsep}{4pt}
\begin{tabular}{@{}llrrrl@{}}
\toprule
Configuration / LR & Seed & $C_{1\times}$ & $C_{4\times}$ & $\Delta C$ & 95\% interval for $\Delta C$ \\
\midrule
144M, selected & 1201 & $-0.020240$ & $+0.003377$ & $+0.023617$ & $[+0.022892,+0.024340]$ \\
144M, selected & 1202 & $-0.015754$ & $+0.002236$ & $+0.017990$ & $[+0.017253,+0.018735]$ \\
388M, 0.004 & 1204 & $-0.029446$ & $-0.005610$ & $+0.023837$ & $[+0.023152,+0.024522]$ \\
388M, 0.004 & 1205 & $-0.033168$ & $-0.006250$ & $+0.026918$ & $[+0.026247,+0.027601]$ \\
\bottomrule
\end{tabular}
\caption{Budget changes after learning-rate selection and in the second configuration. $C=L_{4+}-L_8$ is the test NLL difference in nats/token, and $\Delta C=C_{4\times}-C_{1\times}$. Positive $\Delta C$ denotes contraction, not necessarily reversal. The complete common-rate grid remains in Table~\ref{tab:robust-test}. Intervals are marginal 95\% paired test-block intervals conditional on checkpoints. The 144M rates are selected separately for each window and budget; 388M uses the common coarsely calibrated rate 0.004.}
\label{tab:budget-confirmation}
\end{table*}

\begin{figure*}[t]
\centering\includegraphics[width=\textwidth]{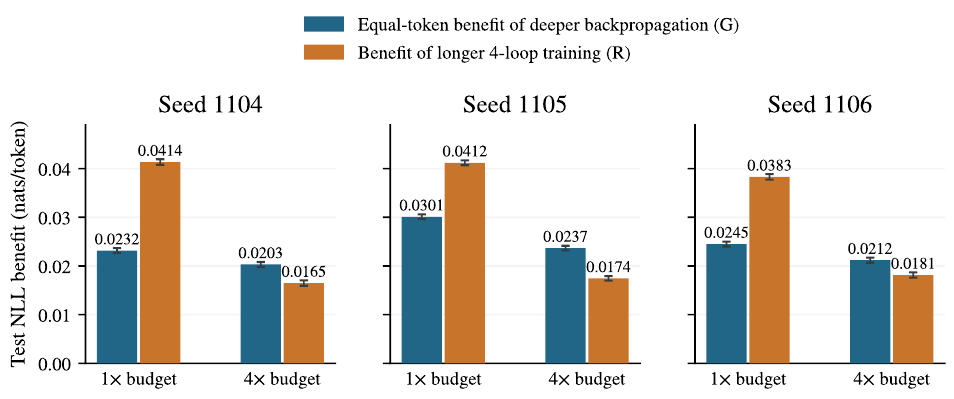}
\caption{\textbf{Extended training gains decrease more than per-token efficiency gains from deeper backpropagation.} Test comparisons use three seeds at common learning rate 0.008 and T16. $L_4$, $L_8$, and $L_{4+}$ denote test NLL for four-iteration backpropagation (K4), eight-iteration backpropagation (K8), and extended four-iteration training (K4+) respectively. Blue bars show $G=L_4-L_8$; orange bars show $R=L_4-L_{4+}$. At $1\times$, extending four-iteration training more than offsets the per-token efficiency gap. At $4\times$, both gains are smaller, and $R$ falls below $G$. The larger decline in $R$ raises $C=G-R$; the smaller value of $G$ partly offsets that change. Error bars show marginal 95\% paired test-block intervals conditional on checkpoints. The controls do not isolate extra tokens from the full optimization schedule and do not decompose the selected-rate or 388M comparisons. Full dev/test values and budget-change intervals appear in Appendix~\ref{app:gr-controls}.}
\label{fig:gr-budget-change}
\end{figure*}

\subsection{Per-token efficiency and extended training}
Let $L_{k,s}^{\mathcal E}(N)$ be endpoint NLL on split $\mathcal E$ for seed $s$ and window $k$, trained from scratch to horizon $N$. For the 144M common-rate controls, architecture, forward depths, inference depth, and base rate 0.008 are fixed. Omitting the common split, define
\begin{align}
G_s(N)&=L_{4,s}(N)-L_{8,s}(N),\label{eq:credit-gap}\\
R_s(N)&=L_{4,s}(N)-L_{4,s}(N^+),\label{eq:extension-gain}\\
C_s(N)&=L_{4,s}(N^+)-L_{8,s}(N)\nonumber\\
&=G_s(N)-R_s(N),\label{eq:allocation-difference}
\end{align}
where $N^+$ is the time-matched extended horizon. $G$ measures the per-token efficiency advantage of K8 over K4, $R$ measures the gain from extending K4, and negative $C$ favors K4+. Hence
\begin{equation}
\Delta C_s=\Delta G_s-\Delta R_s
\label{eq:decomposition}
\end{equation}
for large minus small budget. Each term uses a separately trained endpoint with its own complete schedule. Intermediate checkpoints in longer runs cannot substitute for the equal-token controls.

The completed test controls for seeds 1104/1105/1106 show that $G$ declines by 0.00284--0.00642 nats/token, while $R$ declines by 0.02016--0.02487 (Figure~\ref{fig:gr-budget-change}). The larger reduction in $R$ accounts for positive $\Delta C$ in every seed. For seed 1105, $\Delta G=-0.00642$ and $\Delta R=-0.02376$, yielding $\Delta C=+0.01734$. All three marginal paired intervals place both component changes below zero and the net change above zero.

\paragraph{Relation to diminishing returns.}
A smaller extension benefit can arise from ordinary diminishing returns. For illustration, suppose a fixed-recipe endpoint curve follows $f(D)=L_{\infty}+aD^{-\beta}$, with $a,\beta>0$. Extending its token budget by a fixed factor $q>1$ gives
\begin{equation*}
f(D)-f(qD)=a(1-q^{-\beta})D^{-\beta},
\end{equation*}
which decreases with $D$. This assumed curve is not fitted to our results. It also cannot determine the matched-time ordering without the second window. The additional equal-token runs supply that comparison. The measured window gap decreases too, but by less than the extension benefit. An increasing per-token efficiency gap between K4 and K8 therefore cannot explain contraction in these controls. Each endpoint uses a complete horizon-dependent recipe, so the comparison does not isolate a data-only effect or an optimization mechanism. Two main budgets and one single-seed intermediate point do not determine a crossover location or the shape of either learning curve.

\subsection{Window selection at target budget}
We compare window selection at each budget with transferring the small-budget window, using two new confirmation seeds 1201/1202. Both approaches choose K4+ with rate 0.016 at the small budget. At the large budget, target-budget selection chooses K8 and transfer retains K4+; both use the separately selected large-budget rate 0.008. Transferring the small-budget choice instead of reselecting at the large budget raises test NLL by 0.003377 and 0.002236, with conditional 95\% intervals $[0.002831,0.003917]$ and $[0.001770,0.002704]$ (Table~\ref{tab:choice}). This comparison demonstrates measurable cost of budget-dependent window choice. The comparison uses large-budget calibration and reuses the confirmation endpoints, rather than establishing a new selection algorithm or a saving in search cost. The 24 exploratory endpoints account for approximately 311.8 training GPU-hours, including reused runs. This calibration cost is separate from training the selected recipe.

\begin{table*}[t]
\renewcommand{\arraystretch}{1.4}
\centering\papertable{\begin{tabular}{@{}llrrrrr@{}}
\toprule
Seed & Window & Updates & Hours & Test NLL & CORE $\times100$ & LAMBADA (\%) \\
\midrule
1102 & K4 & 4,986 & 5.2116 & 3.247491 & 10.109 & 22.550 \\
1102 & K6 & 4,498 & 5.2057 & 3.252706 & 10.523 & 23.384 \\
1102 & K8 & 4,096 & 5.1972 & 3.267461 & 10.077 & 23.055 \\
1103 & K4 & 4,986 & 5.2120 & 3.255530 & 9.661 & 22.550 \\
1103 & K6 & 4,498 & 5.1973 & 3.255306 & 9.084 & 22.065 \\
1103 & K8 & 4,096 & 5.2061 & 3.270536 & 10.110 & 22.608 \\
1104 & K4 & 4,986 & 5.2068 & 3.253823 & 10.679 & 22.938 \\
1104 & K6 & 4,498 & 5.1944 & 3.257373 & 10.433 & 22.414 \\
1104 & K8 & 4,096 & 5.2040 & 3.272027 & 11.029 & 23.501 \\
\bottomrule
\end{tabular}}
\caption{Three backward windows at approximately matched small-budget training time. K4+ (extended four-iteration training, 4,986 updates), K6 (six-iteration backpropagation, 4,498 updates), and K8 (eight-iteration backpropagation, 4,096 updates) are compared. All nine endpoints use the 144M configuration, common rate 0.008, and 16 inference iterations. Hours are measured training time. Quality entries are individual-run point estimates; lower test NLL and higher task scores are better. CORE$\times100$ is a chance-adjusted score, not percentage accuracy. LAMBADA uses teacher-forced final-word correctness. Paired task-score intervals remain in Table~\ref{tab:windowpairs}. The comparison does not determine an optimal window at larger budgets.}
\label{tab:windows}
\end{table*}

\paragraph{Training memory and the size of the gain.}
Transferring the small-budget window instead of reselecting at the large budget raises perplexity by about 0.338\% and 0.224\% in seeds 1201/1202, computed as $100(\exp(\Delta \text{NLL})-1)$ where $\Delta \text{NLL}$ is the transfer penalty. These percentages describe relative perplexity changes, not accuracy percentage points. In the recorded 144M implementation, K4 and K8 peak allocated memory are approximately 10.47 and 14.07 GiB. K8 uses about 3.60 GiB more, or 34\% relative to K4. Allocated memory is not the full device-memory requirement; alternative activation-checkpointing and batch configurations were not compared. An NLL preference at approximately matched training time therefore does not establish a joint optimum over loss, memory, and task performance. The memory difference concerns training, so window selection also depends on the available training resources.

\section{Analysis}
\label{sec:downstream}
\subsection{An intermediate backward window}
The K4/K8 comparison does not imply that either is the best window among a wider set of candidates. Table~\ref{tab:windows} brings the six-iteration window into the small-budget comparison; the K1/K2 candidates remain in Table~\ref{tab:grid}. The measured-time horizons are 4,986, 4,498, and 4,096 updates for K4, K6, and K8. Each endpoint is trained from initialization with its own complete schedule.

K6 has lower test NLL than K8 in all three seeds. In seed 1102 it also has higher CORE and LAMBADA scores. That three-metric improvement does not repeat in seeds 1103 or 1104. Compared with K4+, K6 has slightly lower NLL in seed 1103, by about 0.000224 nats/token, but lower task scores. The intermediate window changes the finite-candidate comparison without identifying a consistent multi-metric winner. Pairwise task-score intervals over the common three seeds include zero (Table~\ref{tab:windowpairs}); this does not establish equivalence. The grid covers the small budget and does not determine whether K6 or K8 is preferable at the larger budget.

\subsection{Downstream scope}
Downstream task performance does not follow a uniform pattern. At the small budget, the exploratory LAMBADA mean favors K8 by 0.375 percentage points, whereas both replication seeds favor extended four-iteration training, averaging $-0.669$ points. At the large budget all four accuracy point estimates favor K8, but three individual source intervals cross zero. The exploratory large-budget CORE difference, $+2.246$, becomes $+0.254$ in the replication cohort, with mixed individual signs (Table~\ref{tab:cohorts}).

Small net differences can conceal offsetting output changes. Short-budget seed 1103 has 280 LAMBADA examples correct only for K8 and 277 correct only for K4+. Across six exploratory pairs, no per-pair joint rectangle resolves an opposing word-loss/accuracy preference. Appendices~\ref{app:uncertainty} and~\ref{app:downstreamdetail} retain all task families, margins, and finite-window comparisons. The new selected-rate and 388M models add NLL evidence, not new CORE or LAMBADA results.

\FloatBarrier
\section{Conclusion}
The matched-time NLL advantage of shorter backpropagation diminishes between the tested training budgets. This direction persists across the measured common learning rates, after separate dev-based rate selection, and in a second Parcae configuration. The better of the two windows at the larger budget remains configuration-dependent. Additional equal-token controls show that the window gap also decreases, but less than the benefit of extended training. These controls characterize the change without identifying a causal optimization mechanism. The intermediate-window comparison and measured training memory give further reasons to evaluate candidate recipes under their intended constraints. Backward windows should be compared at the intended training budget and resource constraints, rather than selected once from a small-budget ranking.\label{sec:conclusion}\label{maintext-end}

\section*{Limitations}
Our experiments study two Parcae configurations trained on predominantly English web text across the reported training budgets. The comparisons evaluate complete training recipes, with extended runs using additional tokens and optimization schedules defined for their full training horizons. The reported timing and evaluation results describe the measured A800 setting and selected benchmarks, with computational and reproducibility details provided in Appendices~\ref{app:compute} and~\ref{app:software}.

\section*{Ethical Considerations}
The experiments reuse web-text and benchmark resources and do not recruit human participants or commission new human annotation. Training text is English; the evaluation also includes multilingual language-identification inputs. Web-derived data may contain personal information, stereotypes, and harmful content; deduplication does not remove these risks. The study evaluates training allocation and does not establish the safety or suitability of any checkpoint for deployment. Repeated training and evaluation incur computational and potential environmental costs. Appendix~\ref{app:compute} reports the recorded hardware-time budget and its omissions; power and energy were not measured, so we do not estimate carbon emissions. The supplementary evidence contains numerical observations and hashes, not raw web documents or benchmark text.

\begin{samepage}
\paragraph{AI Statement.}
Generative AI assistants were used primarily to polish the manuscript. Their assistance focused on grammar, formatting, and layout checks. The authors retain responsibility for the scientific content and final manuscript.
\end{samepage}

\bibliography{references}

\appendix
\raggedbottom
\section{Reproducibility Details}
\label{app:recipe}
\subsection{Models, Datasets, and Licenses}
\label{app:assets}
\paragraph{Model and initialization.}
The primary configuration is Parcae-small-140m with 144,323,136 trainable parameters, width 768, feed-forward width 3,072, six attention heads, and a 32,768-token vocabulary. It has two Prelude blocks, two shared Core blocks, and two Coda blocks. Evaluation at 16 recurrences executes 36 Transformer-block applications while storing six unique blocks. The second configuration has 388,003,328 parameters, width 1,024, and four blocks in each component, executing 72 block applications at the same inference recurrence. Both keep all original parameters trainable. Quality models start from seed-specific random canonical initializations rather than pretrained checkpoints. Architecture and native optimizer implementation follow Parcae \citep{prairie2026parcae}. Table~\ref{tab:training} records shared settings; the historical ledger and Tables~\ref{tab:final-endpoints1}--\ref{tab:final-endpoints3} retain per-run horizons, seeds, loss, hours, and allocated memory.

\paragraph{Data provenance.}
The training pool contains 4,979,100 documents and 5,088,221,980 tokens from the first seven shards (000--006) of FineWeb-Edu's \texttt{sample/10BT} configuration. The dataset revision begins \texttt{87f09149ef47}. The tokenizer is \texttt{SandyResearch/parcae-tokenizer}, revision beginning \texttt{6247b5d05928}. Full revision identifiers, source-code fingerprints, and the frozen model configuration are included in the numerical evidence package.

All documents whose hashes occurred in older project data are excluded before assigning the new split. Exact duplicate document hashes are removed across the seven shards. For the retained documents, the first eight SHA-256 bytes, interpreted big-endian modulo 10,000, allocate residues 0--4 and 5--9 to two calibration pools, 10--49 to test, and 50--9,999 to train. The development pool is an older, hash-excluded pool, not a new residue range. Independent split streams concatenate documents with EOS and yield 1,024-target windows with the one-token input/target shift. The stored test pool has 20,367,510 tokens in 19,928 documents; the fixed evaluation uses 10,485,760 target tokens. The longest 144M run consumes 2,614,099,968 targets; the longest 388M run consumes 2,633,891,840. Neither wraps the training stream.

\paragraph{Language, domain, and evaluation sizes.}
FineWeb-Edu is an educational-content-filtered English web corpus \citep{penedo2024fineweb}. The full dev evaluation contains 1,950,720 targets (1,905 windows); monitoring uses 262,144 targets (256 windows). The locked-test campaign evaluates 10,485,760 targets (10,240 windows) per eligible model. These are counts actually consumed, distinct from the stored pool sizes above. Independent English LAMBADA uses all 5,153 evaluation examples \citep{paperno2016lambada}. Table~\ref{tab:tasks} reports every CORE task's evaluated sample count. We do not train or fine-tune on the downstream benchmarks; task-specific in-context demonstrations and the evaluated queries follow the frozen bundle configuration. These are evaluation subsets, not additional pretraining train/dev/test partitions.

The downstream suite spans commonsense reasoning, world knowledge, reading comprehension, language understanding, and symbolic tasks. Prompts are predominantly English, but the language-identification task contains multilingual inputs and symbolic tasks include non-natural-language strings. This is not a representative multilingual or demographic evaluation.

\paragraph{Benchmark attribution.}
CORE uses the public evaluation bundle referenced by Parcae, including its Jeopardy collection.\footnote{\url{https://karpathy-public.s3.us-west-2.amazonaws.com/eval_bundle.zip}. We retain the frozen task configuration and input-file hashes; the bundle is not a new dataset collected for this study.}
Its constituent resources include HellaSwag \citep{zellers-etal-2019-hellaswag}, ARC \citep{clark2018arc}, COPA \citep{roemmele2011copa}, CommonsenseQA \citep{talmor-etal-2019-commonsenseqa}, PIQA \citep{bisk2019piqa}, OpenBookQA \citep{mihaylov2018openbook}, LAMBADA \citep{paperno2016lambada}, Winograd schemas \citep{levesque2012winograd}, WinoGrande \citep{sakaguchi2019winogrande}, BIG-bench tasks \citep{srivastava2023beyond}, the LSAT analytical-reasoning subset of AGIEval \citep{zhong2023agieval}, SQuAD \citep{rajpurkar-etal-2016-squad}, CoQA \citep{reddy-etal-2019-coqa}, and BoolQ \citep{clark-etal-2019-boolq}. Two CORE tasks use the same HellaSwag examples with different shot counts.

\paragraph{Licenses and asset boundaries.}
Parcae's code is MIT-licensed; the FineWeb-Edu release specifies ODC-By.\footnote{\url{https://github.com/sandyresearch/parcae}; \url{https://huggingface.co/datasets/HuggingFaceFW/fineweb-edu}.} The independent LAMBADA input is the \texttt{EleutherAI/lambada\_openai} English release, revision beginning \texttt{900124bf3b82}. Its card carries MIT metadata, while the original LAMBADA distribution specifies CC-BY-4.0.\footnote{\url{https://huggingface.co/datasets/EleutherAI/lambada_openai}; \url{https://huggingface.co/datasets/cimec/lambada}.} These labels concern different assets and do not imply blanket rights to all underlying text. BIG-bench's repository specifies Apache-2.0; component sources and third-party content still require separate attribution.\footnote{\url{https://github.com/google/BIG-bench}.}

The tokenizer revision has no license-bearing model card in the retrieved metadata. We also do not have a complete, independently verified per-component license inventory for the frozen CORE bundle, including the Jeopardy derivative. We therefore do not assert a common license for all evaluation data or redistribute the raw text. The accompanying package contains numerical observations, configuration metadata, source hashes, and manuscript-generation code; it does not contain pretrained weights or the underlying corpora. These provenance and license gaps are stated rather than inferred from the training-code license.

\subsection{Training Hyperparameters}
\label{app:hyperparameters}
\begin{table*}[t]
\centering\papertable{
\begin{tabular}{@{}p{112pt}p{350pt}@{}}
\toprule
Item & Setting \\
\midrule
Model / parameters & Parcae / 144,323,136 and 388,003,328; random initialization \\
Device / GPUs per trial & NVIDIA A800 80GB / 1; up to 8 independent concurrent trials \\
Quality-training updates & 4,096--20,095 updates; exact configurations in the endpoint ledgers \\
Pilot / quality seeds & 1101 / exploration 1102--1104; replications 1105--1106; selected-rate 1201--1202; 388M 1204--1205 \\
Global batch / accumulation & 128 sequences / 16 microbatches of 8 sequences \\
Context / targets per update & 1,024 / 131,072 \\
Sampled forward depth & $\mu=16$ (main), $\mu=8$ (historical secondary); $T=\max(1,\mathrm{Poisson}(\mu))$ \\
Optimizer & Native Muon + AdamW; global gradient clipping at norm 1 \\
Learning-rate scale & 0.008 original controls; complete 144M grid 0.004/0.008/0.016; selected short/long 0.016/0.008; 388M 0.004. Constant first half, linear cooldown thereafter; no warmup \\
Numerical precision & BF16 autocast; FP32 parameters and optimizer states \\
Recorded GPU-hours & 577.1 pre-strengthening process hours; 624.458 added pure-training hours and 0.044419 engineering wall-hours, reported separately in Appendix A.3 \\
\bottomrule
\end{tabular}}
\caption{Shared training configuration and recorded study-level resource summary. Quality horizons exclude the 1,024-update learning-rate pilots and short engineering checks. There is one GPU per trial; eight is the maximum number of concurrent independent trials. Per-model hours and seeds appear in the full ledger.}
\label{tab:training}
\end{table*}

\paragraph{Numerical execution and optimizer.}
FP32 parameters and optimizer states are combined with BF16 autocast, disabling autocast caching; target cross-entropy is FP32 and evaluation loss sums are FP64. Global gradient norm clipping uses threshold 1. Muon momentum is 0.95 with five Newton--Schulz iterations, shape-dependent learning-rate scaling, its native preconditioner and cautious decay. Auxiliary AdamW uses $(\beta_1,\beta_2)=(0.8,0.95)$, $\epsilon=10^{-10}$, and zero weight decay. For zero-based update $u$ and training horizon $N$, the learning-rate scale and Muon weight decay are
\begin{align}
\eta(u;N)&=\eta_0\min\!\left(1,\frac{N-u}{N/2}\right),\\
\lambda_{\rm Muon}(u;N)&=0.2(1-u/N).
\end{align}
The original common recipe uses $\eta_0=0.008$. The complete 144M grid uses 0.004, 0.008, and 0.016; confirmation uses selected short/long rates 0.016/0.008. The 388M comparison uses 0.004. There is no warmup, and auxiliary AdamW has zero weight decay. Each run uses its own full horizon $N$ from initialization.

\paragraph{Batched backward window.}
For a batch with sampled depths $T_i$, let $T_{\max}=\max_i T_i$. Active recurrences are right-aligned: sequence $i$ updates its recurrent state only when the zero-based recurrence index $r\geq T_{\max}-T_i$. Earlier dense Core calls still execute, but their outputs are masked out. The retained window begins at $b=T_{\max}-\min(k,T_{\max})$. If $T_i<k$, this boundary precedes that sequence's first active call, so all its active recurrences retain gradients. The input representation $e$ remains connected to the Prelude through retained injections.

\medskip
\noindent\begin{minipage}{\columnwidth}
\refstepcounter{algorithm}\label{alg:window}
\hrule\smallskip
\textbf{Algorithm \thealgorithm: Batched implementation of a fixed backward window}
\smallskip\hrule\smallskip
\small
\begin{enumerate}\setlength{\itemsep}{2pt}\setlength{\parskip}{0pt}
\item Compute $e=\operatorname{Prelude}(\operatorname{Embed}(\mathrm{tokens}))$ and initialize recurrent state $x$.
\item Set $T_{\max}=\max_i T_i$ and $b=T_{\max}-\min(k,T_{\max})$; set $b=0$ for full backpropagation.
\item For $r=0,\ldots,T_{\max}-1$:
  \begin{enumerate}\setlength{\itemsep}{2pt}
  \item If $r=b>0$, set $x=\operatorname{stopgrad}(x)$.
  \item Record gradients if and only if $r\geq b$.
  \item Compute $z=\operatorname{Core}(\operatorname{Adapter}(x,e))$.
  \item Set $x_i=z_i$ if $r\geq T_{\max}-T_i$; otherwise retain $x_i$.
  \end{enumerate}
\item Record gradients when computing the token loss through projection, Coda, and the prediction head.
\end{enumerate}
\smallskip\hrule
\end{minipage}
\medskip

The notation K1/K2/K4/K6/K8 in the archived ledger denotes fixed windows $k=1,2,4,6,8$. FULL denotes full backpropagation through every sampled recurrence, rather than a fixed window of 16. Projection, normalization, and attention details are omitted from Algorithm~\ref{alg:window}.

\paragraph{Initialization and restoration.}
Constant-window trials start at cursor zero with the same seed-specific random canonical model and an empty optimizer. Paired plans share data permutations and per-sequence forward-depth draws, including identical prefixes when a longer plan is deterministically extended. Inherited quarter-three/four and half-run trials load the complete step-2,048 parent. Their reported recipe costs include the reused prefix; the new work is the suffix only. Constant-window and switching continuation gates compare uninterrupted versus split/restored trajectories. The step-2,048 to step-2,064 validation checks original trace observations and complete legacy/new continuation states; no unavailable historical step-2,064 checkpoint is claimed. All final quality runs have zero skipped updates.

\paragraph{Selection and failures.}
The common 0.008 learning rate comes from the seed-1101 pilot in Section~\ref{sec:design}. The separate main-horizon grid and confirmation are complete (Appendix~\ref{app:lr-snapshot}). Horizon selection uses controlled cost measurements, not dev quality. The two-quarter conditional trigger and timing residual threshold were frozen before the corresponding intervention results. Engineering failures and prelaunch defects are retained in the experiment archives; they are not silently replaced with successful quality trials. An earlier maximum-depth FULL test at microbatch 16 exceeded memory; all compared quality training uses the common validated microbatch eight. This is not an OOM observation from a completed quality arm.

The pilot comprises nine 1,024-update runs: three learning-rate scales for each of four-loop, full-backpropagation, and historical MLMC configurations. The common scale is selected using only the four-loop and full-backpropagation mean endpoint dev loss. It is not a separate hyperparameter optimum for each window. All reported quality endpoints are the prescribed terminal checkpoints, not the best checkpoint along a dev trajectory.

\subsection{Computational Resources}
\label{app:compute}
Experiments run on eight NVIDIA A800 80GB GPUs as independent single-GPU workers. Each training or evaluation job uses one GPU; we sum worker durations rather than multiplying elapsed study time by eight. Tables~\ref{tab:ledger1}--\ref{tab:ledger2} report per-model training hours. The pre-strengthening recorded process subtotal is approximately \textbf{577.1 GPU-hours}, including learning-rate pilots, quality and ablation runs, controlled timing and restoration checks, and the final evaluation campaign (Table~\ref{tab:compute}).

\begin{table*}[t]
\centering\papertable{\begin{tabular}{@{}lrr@{}}
\toprule
Recorded phase & Records & GPU-hours \\
\midrule
Initial pilots, references, and checks & 50 & 42.18 \\
Window/estimator controls and dev evaluation & 20 & 39.92 \\
Half-run interventions and checks & 16 & 9.98 \\
Quarter/short-budget trials and checks & 30 & 62.60 \\
Window/depth/budget trials and checks & 60 & 129.17 \\
Budget replications and checks & 27 & 90.68 \\
Original curves, CORE, LAMBADA, and test & 156 & 77.54 \\
Revision training, checks, and evaluation & 54 & 125.08 \\
Pre-strengthening subtotal & 413 & 577.15 \\
\bottomrule
\end{tabular}}
\caption{Pre-strengthening recorded A800 GPU-time accounting. A record is a worker/job or an early result-level timing interval, not an independent training replication. These categories are disjoint. Prefix reuse is counted only when the prefix is first executed. The total is an approximate recorded subtotal with the omissions described in Appendix~\ref{app:compute}.}
\label{tab:compute}
\end{table*}

This is not a complete measurement of every project activity. Early records measure in-process elapsed time, whereas later records measure worker lifetime, including loading, hashing, CPU/data work, evaluation, and checkpoint overhead. The window/estimator phase uses midpoints of process-lifetime bounds from five-second polling. Early process startup, some failed engineering/OOM attempts, and separate preflight or inference probes lack comparable complete timing coverage and are excluded. CPU-only preparation, audits, manuscript work, idle reservations, and post-experiment occupancy processes are also excluded. Thus 577.1 hours is an approximate subtotal of recorded experiment activity, with incomplete accounting of total project compute; it is not GPU-kernel time, FLOPs, or energy.

The 277.652 hours for new P1--P3 training and the 77.544 hours for the final evaluation workers, reported in Appendix~\ref{app:costs}, are components with narrower accounting scopes, not additional costs to add to Table~\ref{tab:compute}. Reused parent trajectories are charged once in the study total; full-recipe comparisons still include their prefixes. The historical \texttt{submission\_compute.json} preserves 359 intervals totaling 452.0679 hours. The separate revision audit adds 54 disjoint GPU-process intervals totaling 125.0794 hours, giving 577.1473 recorded hours across 413 intervals. Within the revision, nine runs' pure-training time totals 109.3229 hours; it is a component of those process hours, not an additional cost. The extension processes 12,493,520,896 new training targets. CPU audit/statistics wall times are recorded separately and are not CPU-core hours or GPU-hours. The earlier audits, inference probes, and engineering omissions remain unchanged. Revision timing and provenance are preserved alongside the original ledger.

The completed 36-run strengthening experiment adds \textbf{624.458 pure-training GPU-hours}. This is the sum of successful training updates, excluding evaluation, checkpointing, startup, pause writes, and waiting. Its process accounting separately records 626.955 formal-training worker hours, including resumed workers, and 2.220 test worker hours. Reboot recovery splits eight recipes across workers; 44 formal worker records still represent 36 quality runs. Engineering restoration, numerical checks, timing, and stability jobs remain separately itemized in the accompanying audit. Checkpoint and final statistics audits passed.

The frozen window-choice supplement adds no scientific training and opens no new test campaign. Its engineering smoke consumes 0.044419 single-GPU wall-hours. The 24 exploratory endpoints used for calibration total 311.782472 training GPU-hours, comprising 62.321103 at short budget and 249.461369 at long budget. These totals include eight reused endpoints and overlap the experiment ledger; they are not incremental costs to add again. We keep the 624.458-hour pure-training total separate from the earlier 577.1473-hour process subtotal because their scopes differ. Neither subtotal is a complete project-wide measurement of GPU kernels or energy.

\subsection{Software and Evaluation Details}
\label{app:software}
The recorded environment uses Python 3.11.14, PyTorch 2.10.0+cu128, CUDA 12.8, NumPy 1.26.4, Transformers 4.57.1, Tokenizers 0.22.2, Datasets 3.6.0, Jinja2 3.1.6, and PyYAML 6.0.3. Numerical execution uses deterministic algorithms and a fixed cuBLAS workspace configuration. Package repository links and license-check metadata are listed in the supplementary \texttt{asset\_documentation.json}. Dataset/tokenizer revisions and numerical-source fingerprints are preserved in \texttt{training\_recipe.json}; the final evaluation protocol and input hashes are frozen before scoring. The scoring adaptations, prompt construction, truncation, and aggregation are detailed in Appendix~\ref{app:eval}.

\begin{table*}[t]
\centering\papertable{
\begin{tabular}{@{}lll@{}}
\toprule
Comparison & Dev seeds & Test seeds \\
\midrule
Short equal-token K4/K8 & 1102--1106 & 1102--1106 \\
Long equal-token K4/K8 & 1104, 1105, 1106 & 1104, 1105, 1106 \\
Short matched-time K4+/K8 & 1102--1106 & 1102--1106 \\
Long matched-time K4+/K8 & 1102, 1104, 1105, 1106 & 1102, 1104, 1105, 1106 \\
Medium matched-time K4+/K8 & 1104 & 1104 \\
Complete short/long $G/R/C$ & 1104, 1105, 1106 & 1104, 1105, 1106 \\
Short matched-time K4+/K6/K8 & 1102, 1103, 1104 & 1102, 1103, 1104 \\
Full 24-cell LR grid & 1102, 1104 & 1102, 1104 \\
Selected-LR confirmation & 1201, 1202 & 1201, 1202 \\
388M configuration & 1204, 1205 & 1204, 1205 \\
\bottomrule
\end{tabular}}
\caption{Completed dev/test coverage. Original controls and matched-time replications use common rate 0.008. The full grid, selected-rate confirmation, and 388M comparison are listed separately. K4+ denotes extended training. The 36 strengthening endpoints add fixed-T16 NLL evaluation, not downstream or depth-curve coverage.}
\label{tab:coverage}
\end{table*}

\paragraph{Independent runs and reported summaries.}
Every reported configuration--training-seed pair corresponds to one completed quality run. Exploratory seeds 1102/1103/1104 support the short equal-token and matched-time comparisons; 1102/1104 support the long budget and only 1104 the medium budget. Replication seeds 1105/1106 add short and long matched-time pairs. Four subsequent equal-token K4 runs complete their short/long dev and test decompositions. Medium-budget and K6 replications remain outside that addition. Confirmation seeds 1201/1202 follow the complete LR grid, while 1204/1205 supply the 388M comparisons. The newly completed K6/4,498 uses exploratory seed 1104. Native-$\mu=8$ comparisons use 1102/1103. Quarter interventions use 1102/1103; selected half-run replications also use 1104. Tables retain individual values; no mean is a maximum over reruns. Sample SD is reported separately from conditional source-sample intervals, and is not estimated for a single run.

CORE averages prompt seeds 1234, 2345, and 3456, reusing identical zero-shot predictions. Prompt variations are not independent trained models. Independent LAMBADA uses one fixed zero-shot prompt per example. The original FineWeb block analysis uses 20,000 resamples with analysis seed 20261004. Revision source-cluster analysis uses 20,000 resamples with seed 20261005. The completed strengthening dev/test analysis uses 20,000 paired draws with seed 20261006. The three analyses have different sampling units and coverage (Appendix~\ref{app:uncertainty}).

\paragraph{Evidence organization and evaluation history.}
The original final evaluation fixes 61 checkpoint hashes, reproduces their T16 dev endpoints, and retains all 366 dev curve points. Its 47 eligible models receive one test campaign. The nine additional revision models have separately frozen fixed-T16 dev/downstream/test evaluations, with no new depth curves. The 26-model per-example subset comprises 17 old models and those nine additions; it is not the whole historical corpus. Original aggregate scores and test results remain unchanged. Re-collecting old downstream records for margins and pairing is not new independent evidence for their aggregate values.

The completed strengthening set contains 36 new endpoints, comprising four equal-token controls, 16 LR-grid runs, eight selected-rate confirmation runs, and eight 388M runs. All have fixed-T16 dev and test scores. Together with 18 referenced endpoints, the numerical supplement contains 54 complete readouts and 114 paired contrasts. No new CORE, LAMBADA, or six-depth inference scores enter this addition. The earlier model ledger and observations remain unchanged. The supplied final checkpoint and statistics audits both passed. The present manuscript build checks hashes and reconstructs aggregate contrasts locally from those records; it does not repeat checkpoint inference or bootstrap sampling.

\paragraph{Reproducibility package scope.}
The manuscript archive contains saved endpoints, intervals, costs, source manifests, and scripts that regenerate the figures and native tables. A separate completed reproduction archive records ordered patch checks, all 36 recipe/spec matches, and checks of pinned data/tokenizer hashes, packing, and disjoint exact document hashes. Its clean-worktree smoke ran in the existing software environment with width 32 and the real vocabulary. K4 continuous 32 updates agreed with cross-process 16+16 restoration in saved model, optimizer, CPU/CUDA RNG, selection, and cursor state. The smoke also executed K8/16 updates, extended K4/20 updates, T16 evaluation, and paired statistics. The 20-update extension was an engineering check, not a time-matched quality experiment.

The same archive reconstructed 26 historical CORE/LAMBADA model points and 19 paired main intervals/joint rectangles with the original 20,000 source-cluster draws. It reproduced exact McNemar/six-pair Holm, ten additional task, crop, margin, transition, and fixed-seed tables, and three finite-grid tables. All 26 original time/result identities were checked. These are executed reconstructions using the original statistical implementation, not independent algorithmic verification or fresh model inference. Source grouping remains 64,067 CORE clusters and 5,151 LAMBADA paragraph hashes for 5,153 examples, with unknown book identities.

Clean-worktree execution in an existing environment is distinct from a clean installation or a full scientific rerun. The manuscript source archive reproduces numerical displays and document layout; large numeric dependencies and weights remain on the research host. The manifests identify those dependencies but do not constitute a complete standalone redistribution. The accompanying \nolinkurl{credit_depth_software_anonymous.zip} adds portable training, restoration, evaluation, and statistical entry points, 36 strengthening configurations, historical recipe records, and small numerical tables. Package checks reconstruct 114 contrast point estimates from 54 saved endpoints; they do not rerun inference or resampling. Weights, original packed data, exclusion ledgers, and large numeric exports are not included. This is not a full training reproducibility release.

\begin{figure}[t]
\centering\includegraphics[width=\columnwidth]{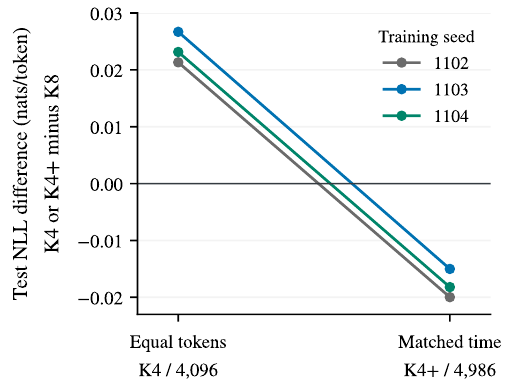}
\caption{The budget definition changes the NLL preference in all three original seeds. Equal-token K4 and K8 train for 4,096 updates; matched-time K4+ uses 4,986. Positive test differences favor K8, negative differences favor the four-loop recipe. Lines connect separate terminal models.}
\label{fig:definition}
\end{figure}

\section{Timing Definitions and Endpoint Checks}
\label{app:timing}
Constant K6/4,096 has lower dev NLL than both four-then-eight and eight-then-four half-run schedules in all three exploratory seeds, by mean gaps of 0.00529 and 0.00679, respectively. Switches at update 2,048 preserve full optimizer/RNG/cursor state and the global LR/WD schedule. Realized switch/K6 training times differ by at most 0.33\% (Table~\ref{tab:k6timing}). This supports only these tested temporal allocations and differs from the K6/4,498 matched-time comparison in the main text. The timing question is related to critical learning periods \citep{achille2019critical}, but does not isolate a biological or schedule-independent critical period.

For constant-window dev endpoints write $G=L_{44}-L_{88}$. For half-run endpoints, $L_{84}$ and $L_{48}$ denote deep-then-shallow and shallow-then-deep. The ordering of the two half-run schedules changes by seed: late deep credit is better in 1102, early deep credit in 1103. Seed 1104's interaction $L_{84}+L_{48}-L_{44}-L_{88}$ is $-0.001059$, or $-4.73\%$ of $G$.

Starting from K4, a single-quarter intervention uses K8 in exactly one 1,024-update quarter. Define $g_q=L_{44}-L_q$ and $w_q=g_q/G$, without clipping or renormalizing. Seed 1102 assigns the largest fractions to quarters three/four (0.319/0.310); 1103 favors quarter one (0.327). No quarter reaches 0.4 in both seeds, so the frozen conditional two-quarter follow-up was not run. These profiles do not identify a common early-credit prescription.

\begin{figure}[t]
\centering\includegraphics[width=\columnwidth]{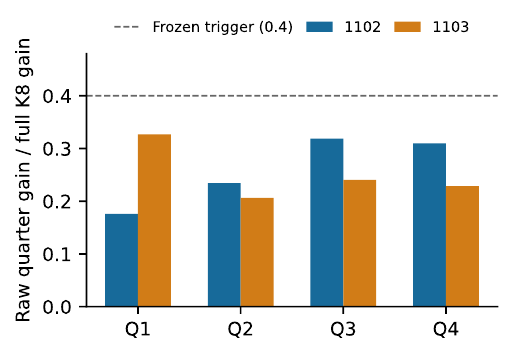}
\caption{Raw single-quarter gains divided by the constant K4--K8 dev gap. Sums are 1.0387 and 1.0028. The frozen 0.4 trigger is not met in both seeds by any quarter.}
\label{fig:quarters}
\end{figure}

Every absolute additivity residual below is less than $0.25G$; the largest is $0.03873G$. This describes the tested endpoints, not arbitrary schedule order invariance. Quarter position changes LR/WD phase and remaining recovery time along with credit timing.

The frozen residuals are
\begin{align}
\epsilon_{\Sigma}&=\sum_{q=1}^4 g_q-G,\\
\epsilon_{84}&=g_1+g_2-(L_{44}-L_{84}),\\
\epsilon_{48}&=g_3+g_4-(L_{44}-L_{48}).
\end{align}
Each check requires $|\epsilon|\leq0.25G$ separately in both original seeds. The threshold is not a confidence interval. The raw ratios in Figure~\ref{fig:quarters} and the residuals in Table~\ref{tab:timing} are computed without clipping or renormalization. Table~\ref{tab:half} reports all half-run endpoints, including the third-seed intervention. $R_{\downarrow}=(L_{44}-L_{84})/G$ and $R_{\uparrow}=(L_{44}-L_{48})/G$ describe fractions of the constant-window gap, not accuracy-retention rates. A majority-retention threshold does not establish quality equivalence.

\section{Training with Mean Forward Depth Eight}
\label{app:mu8}
The secondary setting compares K4/6,494 with full backpropagation at 4,096 updates. Test NLL falls by 0.0798 and 0.0772 nats/token, and FULL minus extended K4 is $-0.618/-0.066$ for CORE$\times100$ and $+0.349/-1.650$ pp for LAMBADA accuracy. Word perplexity improves with extended K4 in both seeds. Reading comprehension and commonsense reasoning improve in both; symbolic problem solving declines in both. These are original point estimates, not covered by the new source-cluster interval subset. Tables~\ref{tab:downstream}, \ref{tab:mu8}, and \ref{tab:mu8taskdeltas} retain the separate FULL reference and mixed task directions. Table~\ref{tab:mu8} reports the secondary comparison. Controlled timing sets the extended \K{4} horizon to 6,494 updates; actual cost ratios relative to FULL remain slightly below one. The two configurations use native $\mu=8$ initialization and a shared paired stream within each seed. All listed test losses use \Te$=16$. The result cannot be transferred to $\mu=16$ as an isolated forward-recurrence effect, or used to choose a universal ratio $k/\mu$.

\section{Downstream Protocol and Aggregation}
\label{app:eval}
The CORE task configuration and baseline metadata are frozen by SHA-256 in the evidence package. Table~\ref{tab:tasks} lists all 22 tasks and their full example counts. For task $t$ and prompt seed $s$, let $a_{ts}$ be its accuracy and $b_t$ its fixed metadata baseline. We compute
\begin{equation}
\mathrm{CORE}_{100}=\frac{100}{3}\sum_s\frac{1}{22}\sum_t\frac{a_{ts}-b_t}{1-b_t}.
\end{equation}
There is no clipping. Scores use prompt seeds 1234, 2345, and 3456; zero-shot instances are evaluated once and their exact scores reused. Thus neither the 66 task/seed cells nor the three prompt seeds should be counted as independent trained models.

CORE inserts BOS=0 and uses EOS=1 and PAD=2. Multiple-choice and schema tasks select the candidate with minimum mean target-token NLL, breaking ties at the first candidate index. Language-modeling tasks require every teacher-forced target-token argmax to match. Candidates are ordered deterministically by length and identifier, right-padded in batches of eight, with shared evaluation RNG settings across models. Left context is cropped to length 1,024 without removing target tokens; 182,665 of the 501,587 candidate sequences per checkpoint require cropping.

For PIQA items 194, 589, and 652, one option is a strict prefix of another. A common-prefix scoring boundary would produce an empty target for an option. We instead score \emph{all} options of each affected item from the question/choice boundary, affecting nine prompt-seed instances in total, and omit none. This is explicitly an adapted CORE implementation. It is not asserted to reproduce a canonical leaderboard. The frozen configuration, baselines, and per-task prompt-seed counts accompany the paper.

Independent LAMBADA uses the full 5,153-example English OpenAI version, separates context and target at the final space, and adds no BOS. Correctness requires all final-word target tokens to match their teacher-forced argmax predictions. Word NLL sums the target-token losses within an example, averages over examples, and exponentiates to obtain word perplexity; it is not token perplexity. These independent scores must not be conflated with CORE's internal BOS-bearing LAMBADA task.

\paragraph{Descriptive task-family decomposition.}
Families follow the dataset directories in the frozen CORE configuration: language understanding (six tasks), world knowledge (four), commonsense reasoning (four), symbolic problem solving (five), and reading comprehension (three). We first average each task's centered score over prompt seeds and subtract paired models; no example counts reweight that score. Tables~\ref{tab:taskdeltas}--\ref{tab:mu8taskdeltas} retain all individual differences. Multiplying a family mean by its task count divided by 22 gives its contribution to the original CORE aggregate. Family labels describe the configuration, not isolated cognitive capabilities. The original summaries are preserved; revision Tables~\ref{tab:revisiontasks1}--\ref{tab:revisiontasks4} additionally show every old and new short/long task and family difference. Family intervals and all prompt-level contributions are supplied in CSV. These descriptive marginal intervals are not corrected across tasks or families. At long budget, old reading-comprehension and commonsense gains do not establish a common replication-cohort mechanism: the new reading-comprehension family differences have opposite signs.

Original test evaluation is one fixed 47-arm campaign after successful prerequisite evaluations. A single opening marker binds the frozen arm list; each arm is evaluated once, and scores are released only after all finish. The reported 14 primary contrasts predate score reading. The revision freezes nine additional arms, each evaluated once at T16 on the same 10,485,760 targets; original test inference is not repeated. New downstream inference and old observation re-collection preserve prompts, batch order, tie rules, and scoring. This revision extension follows earlier benchmark analysis, so it is training-seed replication on studied data, not a fresh test distribution. Additional complete tables are descriptive coverage, not test-driven selection. The strengthening experiment separately completes its declared 36-model fixed-T16 dev/test extension on the same targets, after the frozen dev selection. It repeats neither original test inference nor the downstream/depth-curve campaign. The window-choice supplement reuses eight confirmation checkpoints and their released scores, opening no additional test campaign.

\paragraph{Source recovery and crop sensitivity.}
CORE retains 222,475 unique question/prompt instances and 64,067 recovered source clusters. Shared passages in CoQA, SQuAD, BoolQ, and LSAT use original uncombined sources; prompt variants, both HellaSwag tasks, and recovered cross-task duplicates share weights. Independent LAMBADA has 5,151 paragraph-hash clusters for 5,153 examples; book IDs are unavailable. CORE-internal and independent LAMBADA overlap in source paragraphs, so no independent cross-benchmark $p$-value combination is claimed.

Any-candidate cropping affects 74,921 question/prompt instances (33.676\%); 147,554 are uncropped. This question/prompt fraction differs from the candidate-sequence denominator above. Crop contributions retain original task/prompt denominators and sum to CORE, whereas pooled subgroup accuracy has a different weighting. In the old long-budget seeds, uncropped contributions are $+1.543/+1.730$ and cropped contributions $+0.724/+0.495$ CORE$\times100$. The new cropped contributions are $+0.098/-0.185$, with no common sign. These results do not identify cropping as a causal explanation or rescue the unreplicated broad effect. The CSV retains all groups, intervals, and empty-subgroup rules.

\section{Post Hoc Uncertainty and Sensitivity Analyses}
\label{app:uncertainty}
These analyses were added during manuscript revision, after the original evaluation was frozen and completed. The original block analysis uses released test-loss sums. The revision source-cluster analysis uses paired-example records recovered under the unchanged scoring protocol plus the nine new models. No old checkpoint or aggregate is altered; old test inference is not repeated. The separate nine-arm test extension is disclosed in Appendix~\ref{app:eval}.

\paragraph{Paired contiguous-block bootstrap.}
Each test result preserves 1,280 consecutive batch-loss sums, each covering 8,192 target tokens. We verify their offsets and reconstruct the original aggregate exactly, up to floating-point summation tolerance. The primary analysis groups 16 adjacent batches into 80 non-overlapping blocks of 131,072 targets. For each of the six main matched-time pairs, we resample the same block indices for both checkpoints, with 20,000 replicates and analysis RNG seed 20261004, then report the 2.5/97.5 percentiles of the mean paired NLL difference. Blocks of 65,536 and 262,144 targets give a sensitivity analysis (160 and 40 blocks). Every pair's interval retains its point-estimate sign at all three sizes. Full precision and all sensitivity intervals are in the accompanying analysis JSON; Table~\ref{tab:uncertainty} prints the primary intervals.

Blocks, rather than individual tokens, preserve local sequence dependence. They are contiguous portions of the fixed packed stream, not reconstructed documents; dependence can cross boundaries. The intervals address evaluation-block sampling conditional on the trained checkpoints and do not cover training randomness, new sources, or development selection. The nominal 95\% intervals are marginal and not corrected for simultaneous testing. We do not convert their six signs into six independent replications.

\paragraph{Paired-block analysis for the completed strengthening results.}
The final analysis partitions each evaluation stream into contiguous blocks of 64 sequences, or 65,536 targets. Dev has 30 blocks, the last with 50,176 targets; test has 160 full blocks. Analysis seed 20261006 generates 20,000 shared paired weight vectors across endpoints, budgets, and training seeds within each split. Loss is the weighted block-loss sum divided by weighted target count, preserving the actual mass of the partial dev block. Shared weights retain covariance in $G$, $R$, $C$, their changes, and window-transfer effects.

The 2.5/97.5 percentiles are marginal 95\% intervals conditional on checkpoints and the approximate block unit. Document and book identities are not reconstructed from packed sequences. The intervals neither cover training-seed population variation nor give simultaneous coverage across contrasts. They remain separate from the earlier 131,072-target test-block and downstream source-cluster analyses. The supplied final statistics audit validates all 114 contrasts. The manuscript build independently checks the saved coefficient arithmetic, but copies the reported intervals without drawing new bootstrap samples.

\paragraph{Source-cluster downstream bootstrap.}
The supplied revision uses 20,000 common source-resampling weight vectors, seed 20261005. All predictions from the same recovered source share a multiplicity across prompts, models, and budgets. CORE recomputes each task/prompt denominator per replicate, then retains the original equal-task weighting and chance adjustment. No empty CORE draw occurred; no draws, tasks, or unfavorable instances were discarded. LAMBADA sums word losses and correctness with paragraph weights. Budget changes and fixed-seed averages use these same source weights, preserving covariance. The two benchmark estimators retain their own frozen manifests; their overlapping LAMBADA sources are not treated as independent confirmations.

The 2.5/97.5 percentiles are marginal 95\% source-sample intervals, conditional on checkpoints and the recovered dependence structure. They are not intervals over training seeds, nor proofs of equivalence if they cross zero. Fixed-cohort means and sample SD across seeds remain distinct (Table~\ref{tab:seedsummary}); pooling is descriptive. Unknown book-level and other unrecovered source dependencies remain limitations.

\paragraph{Joint LAMBADA comparisons and discordance.}
For each of six old primary pairs, the joint rectangle uses two 97.5\% marginal intervals (1.25/98.75 percentiles), giving an approximate 95\% per-pair Bonferroni rectangle. Figure~\ref{fig:joint} retains all six pairs. More conservative six-pair familywise marginal bounds and every quadrant fraction are supplied in the CSV. Quadrant bootstrap fractions describe conditional resampling stability, not posterior probabilities or $p$-values. Per-pair coverage must not be presented as simultaneous coverage across all six contrasts.

Exact McNemar tests use the two discordant counts and a two-sided binomial null, with Holm correction only over the six old pairs (Table~\ref{tab:discordance}). The independent-example assumption makes these sensitivity tests when source duplicates exist. No six-pair correction is transferred to newly added contrasts. Correctness counts sum to 5,153 and their difference exactly reconstructs the accuracy delta. All four word-NLL contributions use the full 5,153 denominator (Table~\ref{tab:contributions}); dividing by a near-zero total to create percentage contributions would be unstable.

\paragraph{Tokenization and margins.}
There are 3,099 single-token and 2,054 multi-token LAMBADA targets. Their contributions retain the original full denominator and shared source weights; group signs vary across pairs. The short-budget 1103 single/multi-token NLL contributions are $+0.001808/-0.054459$, while accuracy contributions are $+0.3105/-0.2523$ pp. Seed 1104's short-budget accuracy gap is mainly multi-token; long-budget 1102 improves both strata under K8. These are sensitivity decompositions, not a universal tokenization cause.

For each target position the margin is target logit minus maximum non-target logit. Original argmax tie rules are retained. Figures~\ref{fig:marginfirst}--\ref{fig:marginmin} show first-token and minimum-target-token ECDFs for all six old pairs; group summaries, quantiles, and tie counts remain in the numerical supplement. Every position uses the gold prefix; different competing tokens need not form one word. Hence these are not word-sequence margins or free-generation diagnostics. Minimum-margin changes on jointly wrong examples (K8 minus K4+) are $+0.0251$, $-0.0716$, $-0.0487$, $-0.0644$, $+0.0822$, and $-0.0187$ in the six-pair order. Their signs vary. Separation in uniquely correct groups is partly mechanical from the group definition. Calibration was not measured.

\paragraph{Historical task-omission check.}
The original leave-one-task-out analysis averages the other 21 chance-adjusted task differences, preserving prompt means. K8 remains ahead for every omission in five of six old pairs; short-budget 1102 has both signs. This is an influence diagnostic, not an interval and not a substitute for the new source-cluster analysis. The primary metric always retains all 22 tasks.

\section{Finite Window Frontier and Additional Loss Controls}
\label{app:frontier}
\subsection{Additional endpoint controls}
\label{app:gr-controls}
The complete dev/test decomposition uses seeds 1104/1105/1106 at common learning rate 0.008. Each K4 control starts from its paired canonical state with empty optimizer and cursor zero, and uses a schedule set for the prescribed equal-token horizon. The four added K4 runs complete existing matched-time comparisons, rather than supplying additional independent matched-time replications. Tables~\ref{tab:gr-dev}--\ref{tab:gr-delta-test} retain both splits and conditional budget-change intervals. All underlying endpoints appear in Tables~\ref{tab:final-endpoints1}--\ref{tab:final-endpoints3}.

Both components decline on both splits. On test, $G$ decreases by 0.00284--0.00642 and $R$ by 0.02016--0.02487 nats/token; on dev the ranges are 0.0022--0.0054 and 0.0193--0.0245. The controls do not decompose the separately selected-rate or 388M recipes. No additional seed-1102 long K4 control or 16 matched-state continuation branches were resource-admitted. Their absence is a scheduling decision, not an observed intervention failure.

\begin{table*}[t]
\centering\papertable{
\begin{tabular}{@{}lrrrr@{}}
\toprule
Seed & Budget & $G$ & $R$ & $C=G-R$ \\
\midrule
1104 & 1$\times$ & +0.02239254 & +0.04094663 & -0.01855409 \\
1104 & 4$\times$ & +0.02022664 & +0.01647996 & +0.00374668 \\
1105 & 1$\times$ & +0.03004412 & +0.04077567 & -0.01073154 \\
1105 & 4$\times$ & +0.02463210 & +0.01796940 & +0.00666270 \\
1106 & 1$\times$ & +0.02491055 & +0.03759403 & -0.01268347 \\
1106 & 4$\times$ & +0.02097066 & +0.01832049 & +0.00265016 \\
\bottomrule
\end{tabular}}
\caption{Complete dev decomposition at common LR 0.008. Units are nats/token. Short/long reference horizons are 4,096/16,384; extended K4 horizons are 4,986/19,944. $G=L_4-L_8$, $R=L_4-L_{4+}$, and $C=G-R$. Full-precision endpoint arithmetic precedes rounding.}
\label{tab:gr-dev}
\end{table*}

\begin{table*}[t]
\centering\papertable{
\begin{tabular}{@{}lrrrr@{}}
\toprule
Seed & Budget & $G$ & $R$ & $C=G-R$ \\
\midrule
1104 & 1$\times$ & +0.02315160 & +0.04135504 & -0.01820345 \\
1104 & 4$\times$ & +0.02031574 & +0.01648963 & +0.00382611 \\
1105 & 1$\times$ & +0.03011512 & +0.04120549 & -0.01109037 \\
1105 & 4$\times$ & +0.02369014 & +0.01744454 & +0.00624561 \\
1106 & 1$\times$ & +0.02451100 & +0.03830841 & -0.01379741 \\
1106 & 4$\times$ & +0.02120212 & +0.01814938 & +0.00305273 \\
\bottomrule
\end{tabular}}
\caption{Complete test decomposition for the same three seeds and recipes as Table~\ref{tab:gr-dev}, using 10,485,760 targets at T16. All values are nats/token; negative $C$ favors extended K4.}
\label{tab:gr-test}
\end{table*}

\begin{table*}[t]
\centering\papertable{
\begin{tabular}{@{}llrl@{}}
\toprule
Seed & Contrast & Estimate & 95\% interval \\
\midrule
1104 & $\Delta G$ & -0.00216590 & [-0.003884, -0.000411] \\
1104 & $\Delta R$ & -0.02446667 & [-0.026155, -0.022811] \\
1104 & $\Delta C$ & +0.02230077 & [+0.020466, +0.024063] \\
1105 & $\Delta G$ & -0.00541203 & [-0.007027, -0.003902] \\
1105 & $\Delta R$ & -0.02280627 & [-0.024468, -0.021095] \\
1105 & $\Delta C$ & +0.01739424 & [+0.015841, +0.019015] \\
1106 & $\Delta G$ & -0.00393990 & [-0.005791, -0.002097] \\
1106 & $\Delta R$ & -0.01927353 & [-0.020681, -0.017946] \\
1106 & $\Delta C$ & +0.01533364 & [+0.013750, +0.017051] \\
\bottomrule
\end{tabular}}
\caption{Dev long-minus-short component changes with marginal 95\% paired-block intervals conditional on checkpoints. All three seeds and both component directions are retained.}
\label{tab:gr-delta-intervals}
\end{table*}

\begin{table*}[t]
\centering\papertable{
\begin{tabular}{@{}llrl@{}}
\toprule
Seed & Contrast & Estimate & 95\% interval \\
\midrule
1104 & $\Delta G$ & -0.00283585 & [-0.003527, -0.002155] \\
1104 & $\Delta R$ & -0.02486541 & [-0.025654, -0.024107] \\
1104 & $\Delta C$ & +0.02202956 & [+0.021298, +0.022777] \\
1105 & $\Delta G$ & -0.00642498 & [-0.007023, -0.005828] \\
1105 & $\Delta R$ & -0.02376095 & [-0.024456, -0.023088] \\
1105 & $\Delta C$ & +0.01733598 & [+0.016598, +0.018077] \\
1106 & $\Delta G$ & -0.00330888 & [-0.003933, -0.002687] \\
1106 & $\Delta R$ & -0.02015903 & [-0.020902, -0.019414] \\
1106 & $\Delta C$ & +0.01685015 & [+0.016121, +0.017585] \\
\bottomrule
\end{tabular}}
\caption{Test long-minus-short component changes. Intervals use the test blocks and are not copied from dev or interpreted as training-seed population uncertainty.}
\label{tab:gr-delta-test}
\end{table*}

\subsection{Completed learning-rate and configuration checks}
\label{app:lr-snapshot}
The full 144M grid contains 24 endpoints from two windows, two budgets, three rates, and exploratory seeds 1102/1104 (Table~\ref{tab:lr-grid}). Eight previous 0.008 endpoints are reused and 16 are new. Each window/budget rate is selected by the two seeds' mean terminal dev NLL. Exact ties prefer 0.008, then 0.004, then 0.016. All 0.016 runs completed. The selected short/long rates are 0.016/0.008 for both windows (Table~\ref{tab:lr-selection}); 0.016 lies at the upper grid boundary. Test values are reported after selection, not used to choose rates.

All eight confirmation endpoints for 1201/1202 completed after selection was frozen. Every common-rate long-minus-short $C$ is positive on both splits, but six rate/seed contrasts share two seeds and several checkpoints. Tables~\ref{tab:robust-dev} and~\ref{tab:robust-test} separate the original-rate grid, selected-rate confirmation, and second configuration. The test sign at the larger budget depends on the configuration and recipe, even when the budget-change sign agrees.

All eight 388M endpoints completed at common LR 0.004. The dedicated seed-1203, 200-update calibration chose that rate before quality training; it does not establish independent full-horizon optima. K8 horizons are 4,096/16,384 and K4+ horizons 5,024/20,095. Width, physical block counts, and tokens per parameter differ from the 144M configuration, so this is a second configuration in the same family rather than an isolated width intervention.

The original final checkpoint audit covers all 36 new runs, their canonical/parent states, optimizer ownership and trainability, paired data/depth/selection/RNG/cursors, schedules, budgets, and zero skipped updates. A separate final statistics audit also passed. All reported endpoints are complete.

\begin{table*}[t]
\centering\papertable{
\begin{tabular}{@{}lllrrr@{}}
\toprule
Seed & Window & Budget & LR & Dev NLL & Test NLL \\
\midrule
1102 & K4+ & 1$\times$ & 0.004 & 3.26053613 & 3.27251953 \\
1102 & K4+ & 1$\times$ & 0.008 & 3.23426446 & 3.24749080 \\
1102 & K4+ & 1$\times$ & 0.016 & 3.23017013 & 3.24195175 \\
1102 & K8 & 1$\times$ & 0.004 & 3.27871977 & 3.29114672 \\
1102 & K8 & 1$\times$ & 0.008 & 3.25436555 & 3.26746083 \\
1102 & K8 & 1$\times$ & 0.016 & 3.25485746 & 3.26725415 \\
1102 & K4+ & 4$\times$ & 0.004 & 3.06107754 & 3.07048581 \\
1102 & K4+ & 4$\times$ & 0.008 & 3.04398375 & 3.05467151 \\
1102 & K4+ & 4$\times$ & 0.016 & 3.06518582 & 3.07474131 \\
1102 & K8 & 4$\times$ & 0.004 & 3.05562497 & 3.06597372 \\
1102 & K8 & 4$\times$ & 0.008 & 3.04575550 & 3.05563557 \\
1102 & K8 & 4$\times$ & 0.016 & 3.06316954 & 3.07298423 \\
1104 & K4+ & 1$\times$ & 0.004 & 3.25948333 & 3.27174232 \\
1104 & K4+ & 1$\times$ & 0.008 & 3.24249376 & 3.25382316 \\
1104 & K4+ & 1$\times$ & 0.016 & 3.23408411 & 3.24604778 \\
1104 & K8 & 1$\times$ & 0.004 & 3.27994775 & 3.29235186 \\
1104 & K8 & 1$\times$ & 0.008 & 3.26104786 & 3.27202661 \\
1104 & K8 & 1$\times$ & 0.016 & 3.25553928 & 3.26798829 \\
1104 & K4+ & 4$\times$ & 0.004 & 3.05405555 & 3.06377673 \\
1104 & K4+ & 4$\times$ & 0.008 & 3.05052683 & 3.06052223 \\
1104 & K4+ & 4$\times$ & 0.016 & 3.06359508 & 3.07379703 \\
1104 & K8 & 4$\times$ & 0.004 & 3.05621534 & 3.06588042 \\
1104 & K8 & 4$\times$ & 0.008 & 3.04678015 & 3.05669612 \\
1104 & K8 & 4$\times$ & 0.016 & 3.06335317 & 3.07274713 \\
\bottomrule
\end{tabular}}
\caption{All 24 144M exploration endpoints. K4+ and K8 approximately match time within seed/budget; both dev and test NLL use T16. Rate selection uses dev means over the two seeds, never the displayed test values.}
\label{tab:lr-grid}
\end{table*}

\begin{table*}[t]
\centering\papertable{
\begin{tabular}{@{}llrrl@{}}
\toprule
Window & Budget & LR & Mean dev NLL & Choice \\
\midrule
K4+ & short & 0.008 & 3.23837911 &  \\
K4+ & short & 0.004 & 3.26000973 &  \\
K4+ & short & 0.016 & 3.23212712 & Selected \\
K8 & short & 0.008 & 3.25770670 &  \\
K8 & short & 0.004 & 3.27933376 &  \\
K8 & short & 0.016 & 3.25519837 & Selected \\
K4+ & long & 0.008 & 3.04725529 & Selected \\
K4+ & long & 0.004 & 3.05756654 &  \\
K4+ & long & 0.016 & 3.06439045 &  \\
K8 & long & 0.008 & 3.04626782 & Selected \\
K8 & long & 0.004 & 3.05592016 &  \\
K8 & long & 0.016 & 3.06326135 &  \\
\bottomrule
\end{tabular}}
\caption{Mean dev NLL over exploration seeds 1102/1104. Rate selection is performed independently for each window/budget using the frozen tie rule. Short-budget optima are at the grid upper boundary.}
\label{tab:lr-selection}
\end{table*}

\begin{table*}[t]
\centering\papertable{
\begin{tabular}{@{}lrrrrl@{}}
\toprule
Configuration / LR & Seed & $C_{1\times}$ & $C_{4\times}$ & $\Delta C$ & 95\% interval for $\Delta C$ \\
\midrule
144M, 0.004 & 1102 & -0.01818363 & +0.00545257 & +0.02363621 & [+0.022229, +0.025012] \\
144M, 0.008 & 1102 & -0.02010109 & -0.00177175 & +0.01832935 & [+0.016633, +0.020051] \\
144M, 0.016 & 1102 & -0.02468733 & +0.00201628 & +0.02670361 & [+0.025086, +0.028266] \\
144M, 0.004 & 1104 & -0.02046442 & -0.00215979 & +0.01830463 & [+0.016461, +0.020211] \\
144M, 0.008 & 1104 & -0.01855409 & +0.00374668 & +0.02230077 & [+0.020466, +0.024063] \\
144M, 0.016 & 1104 & -0.02145517 & +0.00024191 & +0.02169708 & [+0.019410, +0.023881] \\
144M, selected & 1201 & -0.01860997 & +0.00399745 & +0.02260742 & [+0.020974, +0.024360] \\
144M, selected & 1202 & -0.01512557 & +0.00170803 & +0.01683360 & [+0.015358, +0.018323] \\
388M, 0.004 & 1204 & -0.02979654 & -0.00559073 & +0.02420580 & [+0.022540, +0.025928] \\
388M, 0.004 & 1205 & -0.03224246 & -0.00682949 & +0.02541297 & [+0.023931, +0.026838] \\
\bottomrule
\end{tabular}}
\caption{All dev matched-time budget changes from the complete grid, selected-rate confirmation, and 388M configuration. $C=L_{4+}-L_8$; $\Delta C=C_{4\times}-C_{1\times}$. Intervals are conditional paired-block intervals.}
\label{tab:robust-dev}
\end{table*}

\begin{table*}[t]
\centering\papertable{
\begin{tabular}{@{}lrrrrl@{}}
\toprule
Configuration / LR & Seed & $C_{1\times}$ & $C_{4\times}$ & $\Delta C$ & 95\% interval for $\Delta C$ \\
\midrule
144M, 0.004 & 1102 & -0.01862719 & +0.00451209 & +0.02313928 & [+0.022434, +0.023846] \\
144M, 0.008 & 1102 & -0.01997003 & -0.00096407 & +0.01900596 & [+0.018242, +0.019760] \\
144M, 0.016 & 1102 & -0.02530240 & +0.00175708 & +0.02705948 & [+0.026363, +0.027742] \\
144M, 0.004 & 1104 & -0.02060954 & -0.00210369 & +0.01850585 & [+0.017797, +0.019222] \\
144M, 0.008 & 1104 & -0.01820345 & +0.00382611 & +0.02202956 & [+0.021298, +0.022777] \\
144M, 0.016 & 1104 & -0.02194051 & +0.00104990 & +0.02299040 & [+0.022237, +0.023748] \\
144M, selected & 1201 & -0.02023992 & +0.00337700 & +0.02361693 & [+0.022892, +0.024340] \\
144M, selected & 1202 & -0.01575372 & +0.00223639 & +0.01799010 & [+0.017253, +0.018735] \\
388M, 0.004 & 1204 & -0.02944630 & -0.00560962 & +0.02383668 & [+0.023152, +0.024522] \\
388M, 0.004 & 1205 & -0.03316811 & -0.00625027 & +0.02691784 & [+0.026247, +0.027601] \\
\bottomrule
\end{tabular}}
\caption{Corresponding test contrasts in nats/token, retaining all signs. Positive $\Delta C$ means contraction of the K4+ advantage; it does not require positive long-budget $C$. The six grid rows are two seeds measured at three rates.}
\label{tab:robust-test}
\end{table*}

\subsection{Frozen window selection and transfer}
\label{app:choice}
The window-choice rule and its predictions were frozen before target outcomes, as verified by the supplied prospective audit. Exploration means over 1102/1104 first select a rate independently for each window and budget, then select the window, with an exact window tie favoring K4+. Budget-specific selection chooses K4+/0.016 at the short budget and K8/0.008 at the long budget. Transfer keeps the short-budget window but still receives that window's separately selected long-budget rate. The comparison therefore transfers the window, not the learning rate.

The eight confirmation endpoints of 1201/1202 are reused; this analysis adds neither independent training nor a new test campaign. The transfer effect $Q$ is transfer minus budget-specific test NLL. Short-budget $Q=0$ because both rules select the identical checkpoint. At long budget, positive $Q$ favors selection at the target budget (Table~\ref{tab:choice}). Long-budget calibration is required by both rules, so the comparison gives no net search-cost saving or general regret guarantee.

The numerical supplement retains all four rules, including always-K4+ and always-K8, and their recorded costs. Its minimum over two observed checkpoints is a descriptive reference, not a test-based selection procedure. Conditional regret intervals recompute that minimum in each original shared paired draw. They do not imply that a future seed will attain the same optimum.

\begin{table*}[t]
\centering\papertable{
\begin{tabular}{@{}llllrl@{}}
\toprule
Seed & Budget & Target-budget choice & Transferred window & Test penalty & 95\% interval \\
\midrule
1201 & short & K4+, 0.016 & K4+, 0.016 & 0.00000000 & [0.000000, 0.000000] \\
1201 & long & K8, 0.008 & K4+, 0.008 & 0.00337700 & [0.002831, 0.003917] \\
1202 & short & K4+, 0.016 & K4+, 0.016 & 0.00000000 & [0.000000, 0.000000] \\
1202 & long & K8, 0.008 & K4+, 0.008 & 0.00223639 & [0.001770, 0.002704] \\
\bottomrule
\end{tabular}}
\caption{Frozen target-budget selection versus short-window transfer on the same confirmation models. Rates appear beside the chosen windows. Test penalty $Q$ is transfer minus target-budget NLL in nats/token. Exact short-budget zeros reuse identical checkpoints. Brackets give marginal 95\% paired test-block intervals conditional on the checkpoints.}
\label{tab:choice}
\end{table*}

\subsection{Complete endpoint and cost ledger}
\label{app:final-ledger}
Each endpoint below is a prescribed terminal checkpoint, not a minimum selected along a learning curve. F01--F18 are referenced results; F19--F54 are the 36 new scientific runs. IDs map to original run names and source hashes in the numerical supplement. Training targets equal updates times 131,072. Hours are pure training; allocated memory is the full-recipe peak, including both sides of resumed runs. Fixed-T16 dev and test cover every row. The first 18 references are not charged again as newly executed training.

\begin{table*}[t]
\centering\papertable{
\begin{tabular}{@{}llllrrrrrr@{}}
\toprule
ID & Model & Seed & Window & Updates & LR & Dev NLL & Test NLL & Hours & GiB \\
\midrule
F01 & 144M & 1102 & K8 & 4,096 & 0.008 & 3.25436555 & 3.26746083 & 5.197 & 14.07 \\
F02 & 144M & 1102 & K4 & 4,986 & 0.008 & 3.23426446 & 3.24749080 & 5.212 & 10.47 \\
F03 & 144M & 1102 & K8 & 16,384 & 0.008 & 3.04575550 & 3.05563557 & 20.742 & 14.07 \\
F04 & 144M & 1102 & K4 & 19,944 & 0.008 & 3.04398375 & 3.05467151 & 20.773 & 10.47 \\
F05 & 144M & 1104 & K8 & 4,096 & 0.008 & 3.26104786 & 3.27202661 & 5.204 & 14.07 \\
F06 & 144M & 1104 & K4 & 4,986 & 0.008 & 3.24249376 & 3.25382316 & 5.207 & 10.47 \\
F07 & 144M & 1104 & K8 & 16,384 & 0.008 & 3.04678015 & 3.05669612 & 20.789 & 14.07 \\
F08 & 144M & 1104 & K4 & 19,944 & 0.008 & 3.05052683 & 3.06052223 & 20.769 & 10.47 \\
F09 & 144M & 1104 & K4 & 4,096 & 0.008 & 3.28344040 & 3.29517821 & 4.268 & 10.47 \\
F10 & 144M & 1104 & K4 & 16,384 & 0.008 & 3.06700679 & 3.07701187 & 17.075 & 10.47 \\
F11 & 144M & 1105 & K8 & 16,384 & 0.008 & 3.04567223 & 3.05671211 & 20.736 & 14.07 \\
F12 & 144M & 1106 & K8 & 16,384 & 0.008 & 3.04763206 & 3.05747374 & 20.803 & 14.07 \\
F13 & 144M & 1105 & K4 & 19,944 & 0.008 & 3.05233493 & 3.06295772 & 20.889 & 10.47 \\
F14 & 144M & 1106 & K4 & 19,944 & 0.008 & 3.05028222 & 3.06052647 & 20.866 & 10.47 \\
F15 & 144M & 1105 & K8 & 4,096 & 0.008 & 3.25015758 & 3.26282599 & 5.208 & 14.07 \\
F16 & 144M & 1106 & K8 & 4,096 & 0.008 & 3.24943345 & 3.26284120 & 5.201 & 14.07 \\
F17 & 144M & 1105 & K4 & 4,986 & 0.008 & 3.23942604 & 3.25173562 & 5.200 & 10.47 \\
F18 & 144M & 1106 & K4 & 4,986 & 0.008 & 3.23674998 & 3.24904379 & 5.226 & 10.47 \\
\bottomrule
\end{tabular}}
\caption{Complete strengthening endpoint ledger, part 1. NLL is nats/token, hours are pure training, and GiB is peak allocated memory. All original observations are retained. IDs are defined in Appendix~\ref{app:final-ledger} and the numerical supplement.}
\label{tab:final-endpoints1}
\end{table*}

\begin{table*}[t]
\centering\papertable{
\begin{tabular}{@{}llllrrrrrr@{}}
\toprule
ID & Model & Seed & Window & Updates & LR & Dev NLL & Test NLL & Hours & GiB \\
\midrule
F19 & 144M & 1105 & K4 & 16,384 & 0.008 & 3.07030433 & 3.08040226 & 17.162 & 10.47 \\
F20 & 144M & 1106 & K4 & 16,384 & 0.008 & 3.06860272 & 3.07867586 & 17.174 & 10.47 \\
F21 & 144M & 1105 & K4 & 4,096 & 0.008 & 3.28020171 & 3.29294112 & 4.279 & 10.47 \\
F22 & 144M & 1106 & K4 & 4,096 & 0.008 & 3.27434400 & 3.28735220 & 4.278 & 10.47 \\
F23 & 144M & 1102 & K4 & 19,944 & 0.004 & 3.06107754 & 3.07048581 & 20.832 & 10.47 \\
F24 & 144M & 1102 & K8 & 16,384 & 0.004 & 3.05562497 & 3.06597372 & 20.765 & 14.07 \\
F25 & 144M & 1104 & K4 & 19,944 & 0.004 & 3.05405555 & 3.06377673 & 20.843 & 10.47 \\
F26 & 144M & 1104 & K8 & 16,384 & 0.004 & 3.05621534 & 3.06588042 & 20.747 & 14.07 \\
F27 & 144M & 1102 & K4 & 19,944 & 0.016 & 3.06518582 & 3.07474131 & 20.918 & 10.47 \\
F28 & 144M & 1102 & K8 & 16,384 & 0.016 & 3.06316954 & 3.07298423 & 20.641 & 14.07 \\
F29 & 144M & 1104 & K4 & 19,944 & 0.016 & 3.06359508 & 3.07379703 & 20.900 & 10.47 \\
F30 & 144M & 1104 & K8 & 16,384 & 0.016 & 3.06335317 & 3.07274713 & 20.744 & 14.07 \\
F31 & 144M & 1102 & K4 & 4,986 & 0.004 & 3.26053613 & 3.27251953 & 5.199 & 10.47 \\
F32 & 144M & 1102 & K8 & 4,096 & 0.004 & 3.27871977 & 3.29114672 & 5.198 & 14.07 \\
F33 & 144M & 1104 & K4 & 4,986 & 0.004 & 3.25948333 & 3.27174232 & 5.232 & 10.47 \\
F34 & 144M & 1104 & K8 & 4,096 & 0.004 & 3.27994775 & 3.29235186 & 5.161 & 14.07 \\
F35 & 144M & 1102 & K4 & 4,986 & 0.016 & 3.23017013 & 3.24195175 & 5.175 & 10.47 \\
F36 & 144M & 1102 & K8 & 4,096 & 0.016 & 3.25485746 & 3.26725415 & 5.189 & 14.07 \\
\bottomrule
\end{tabular}}
\caption{Complete strengthening endpoint ledger, part 2. NLL is nats/token, hours are pure training, and GiB is peak allocated memory. All original observations are retained. IDs are defined in Appendix~\ref{app:final-ledger} and the numerical supplement.}
\label{tab:final-endpoints2}
\end{table*}

\begin{table*}[t]
\centering\papertable{
\begin{tabular}{@{}llllrrrrrr@{}}
\toprule
ID & Model & Seed & Window & Updates & LR & Dev NLL & Test NLL & Hours & GiB \\
\midrule
F37 & 144M & 1104 & K4 & 4,986 & 0.016 & 3.23408411 & 3.24604778 & 5.190 & 10.47 \\
F38 & 144M & 1104 & K8 & 4,096 & 0.016 & 3.25553928 & 3.26798829 & 5.159 & 14.07 \\
F39 & 388M & 1204 & K8 & 16,384 & 0.004 & 2.87061558 & 2.88054893 & 53.951 & 31.74 \\
F40 & 388M & 1204 & K4 & 20,095 & 0.004 & 2.86502485 & 2.87493930 & 54.174 & 22.34 \\
F41 & 388M & 1205 & K8 & 16,384 & 0.004 & 2.87040951 & 2.87995483 & 53.966 & 31.74 \\
F42 & 388M & 1205 & K4 & 20,095 & 0.004 & 2.86358003 & 2.87370456 & 53.978 & 22.34 \\
F43 & 388M & 1204 & K8 & 4,096 & 0.004 & 3.14724896 & 3.15859669 & 13.455 & 31.74 \\
F44 & 388M & 1204 & K4 & 5,024 & 0.004 & 3.11745242 & 3.12915039 & 13.473 & 22.34 \\
F45 & 388M & 1205 & K8 & 4,096 & 0.004 & 3.14656076 & 3.15820928 & 13.481 & 31.74 \\
F46 & 388M & 1205 & K4 & 5,024 & 0.004 & 3.11431830 & 3.12504117 & 13.445 & 22.34 \\
F47 & 144M & 1201 & K4 & 19,944 & 0.008 & 3.05205213 & 3.06165056 & 20.734 & 10.47 \\
F48 & 144M & 1201 & K8 & 16,384 & 0.008 & 3.04805468 & 3.05827356 & 20.703 & 14.07 \\
F49 & 144M & 1202 & K4 & 19,944 & 0.008 & 3.04854450 & 3.05896458 & 20.778 & 10.47 \\
F50 & 144M & 1202 & K8 & 16,384 & 0.008 & 3.04683646 & 3.05672819 & 20.801 & 14.07 \\
F51 & 144M & 1201 & K4 & 4,986 & 0.016 & 3.23817337 & 3.25003592 & 5.178 & 10.47 \\
F52 & 144M & 1201 & K8 & 4,096 & 0.016 & 3.25678334 & 3.27027584 & 5.178 & 14.07 \\
F53 & 144M & 1202 & K4 & 4,986 & 0.016 & 3.23490333 & 3.24746698 & 5.190 & 10.47 \\
F54 & 144M & 1202 & K8 & 4,096 & 0.016 & 3.25002890 & 3.26322069 & 5.188 & 14.07 \\
\bottomrule
\end{tabular}}
\caption{Complete strengthening endpoint ledger, part 3. NLL is nats/token, hours are pure training, and GiB is peak allocated memory. All original observations are retained. IDs are defined in Appendix~\ref{app:final-ledger} and the numerical supplement.}
\label{tab:final-endpoints3}
\end{table*}

\begin{table*}[t]
\centering\papertable{
\begin{tabular}{@{}lrrrrrrr@{}}
\toprule
K4+ ID & K8 ID & K4+ hours & K8 hours & $\Delta$ time (\%) & Extra targets & K4+ GiB & K8 GiB \\
\midrule
F06 & F05 & 5.2068 & 5.2040 & +0.053 & 116,654,080 & 10.47 & 14.07 \\
F08 & F07 & 20.7687 & 20.7886 & -0.096 & 466,616,320 & 10.47 & 14.07 \\
F17 & F15 & 5.2001 & 5.2078 & -0.147 & 116,654,080 & 10.47 & 14.07 \\
F13 & F11 & 20.8891 & 20.7357 & +0.740 & 466,616,320 & 10.47 & 14.07 \\
F18 & F16 & 5.2260 & 5.2007 & +0.486 & 116,654,080 & 10.47 & 14.07 \\
F14 & F12 & 20.8663 & 20.8029 & +0.305 & 466,616,320 & 10.47 & 14.07 \\
F31 & F32 & 5.1990 & 5.1976 & +0.026 & 116,654,080 & 10.47 & 14.07 \\
F23 & F24 & 20.8324 & 20.7646 & +0.327 & 466,616,320 & 10.47 & 14.07 \\
F02 & F01 & 5.2116 & 5.1972 & +0.277 & 116,654,080 & 10.47 & 14.07 \\
F04 & F03 & 20.7728 & 20.7417 & +0.150 & 466,616,320 & 10.47 & 14.07 \\
F35 & F36 & 5.1751 & 5.1891 & -0.270 & 116,654,080 & 10.47 & 14.07 \\
F27 & F28 & 20.9179 & 20.6406 & +1.343 & 466,616,320 & 10.47 & 14.07 \\
F33 & F34 & 5.2319 & 5.1606 & +1.380 & 116,654,080 & 10.47 & 14.07 \\
\bottomrule
\end{tabular}}
\caption{Realized matched-time comparisons, part 1. IDs identify K4+/K8 endpoints in the preceding ledger. Time mismatch is $100(t_{4+}/t_8-1)$. Extra targets and both memory peaks are reported without post hoc time filtering. Repeated pairs used by different analyses are not additional training runs. GiB denotes peak allocated memory.}
\label{tab:final-costs1}
\end{table*}

\begin{table*}[t]
\centering\papertable{
\begin{tabular}{@{}lrrrrrrr@{}}
\toprule
K4+ ID & K8 ID & K4+ hours & K8 hours & $\Delta$ time (\%) & Extra targets & K4+ GiB & K8 GiB \\
\midrule
F25 & F26 & 20.8428 & 20.7472 & +0.461 & 466,616,320 & 10.47 & 14.07 \\
F06 & F05 & 5.2068 & 5.2040 & +0.053 & 116,654,080 & 10.47 & 14.07 \\
F08 & F07 & 20.7687 & 20.7886 & -0.096 & 466,616,320 & 10.47 & 14.07 \\
F37 & F38 & 5.1897 & 5.1585 & +0.604 & 116,654,080 & 10.47 & 14.07 \\
F29 & F30 & 20.8998 & 20.7444 & +0.749 & 466,616,320 & 10.47 & 14.07 \\
F51 & F52 & 5.1775 & 5.1784 & -0.017 & 116,654,080 & 10.47 & 14.07 \\
F47 & F48 & 20.7343 & 20.7033 & +0.150 & 466,616,320 & 10.47 & 14.07 \\
F53 & F54 & 5.1902 & 5.1880 & +0.044 & 116,654,080 & 10.47 & 14.07 \\
F49 & F50 & 20.7781 & 20.8010 & -0.110 & 466,616,320 & 10.47 & 14.07 \\
F44 & F43 & 13.4726 & 13.4554 & +0.128 & 121,634,816 & 22.34 & 31.74 \\
F40 & F39 & 54.1737 & 53.9515 & +0.412 & 486,408,192 & 22.34 & 31.74 \\
F46 & F45 & 13.4449 & 13.4807 & -0.266 & 121,634,816 & 22.34 & 31.74 \\
F42 & F41 & 53.9777 & 53.9661 & +0.021 & 486,408,192 & 22.34 & 31.74 \\
\bottomrule
\end{tabular}}
\caption{Realized matched-time comparisons, part 2. IDs identify K4+/K8 endpoints in the preceding ledger. Time mismatch is $100(t_{4+}/t_8-1)$. Extra targets and both memory peaks are reported without post hoc time filtering. Repeated pairs used by different analyses are not additional training runs. GiB denotes peak allocated memory.}
\label{tab:final-costs2}
\end{table*}

\subsection{Finite window comparisons}

The short-cost grid allocates 5,910, 5,564, 4,986, 4,498, and 4,096 updates to windows 1, 2, 4, 6, and 8, respectively, based on controlled costs before quality readout. K1 and K2 have higher test NLL than K4 and K6 despite receiving more tokens (Table~\ref{tab:grid}). Dev has a unique minimum at K4 in both original seeds, but its seed-1103 margin over K6 is only 0.000143. Test favors K6 for that seed by 0.000224. This finite grid does not establish a robust unique optimum.

The equal-token grid's raw adjacent dev improvements per unit increase in $k\in\{1,2,4,6,8\}$ decrease in both original seeds, without using a tolerance. FULL is a separate variable-depth endpoint. Figure~\ref{fig:tradeoff} retains the loss-only comparisons; Figure~\ref{fig:long} supplies the corresponding dev/test budget comparisons. All 14 original primary contrasts preserve their dev/test directions, but they overlap in checkpoints and seeds and are not independent confirmations.

\section{Historical Estimator Controls}
\label{app:controls}
The frozen final list retains FULL, an importance-weighted telescoping depth-gradient estimator (MLMC), and its unweighted correction ablation from earlier development. They are not proposed as successful new algorithms in this paper. In the original two-seed 4,096-update comparison, MLMC did not satisfy its predefined improvement gate relative to \K{4}. Their complete endpoint losses, downstream scores, and inference curves remain visible in the model ledger and raw tables. Randomized telescoping can target a full-depth gradient in expectation, but this fact alone does not describe the effects of finite-sample variance, global clipping, nonlinear optimizer updates, or entire learned trajectories \citep{tallec2017unbiasing,beatson2019loops}. A negative result for the tested estimator and public learning-rate choice is not a theorem about unbiased estimators.

\section{Costs and Coverage}
\label{app:costs}
Tables~\ref{tab:ledger1}--\ref{tab:ledger2} list the original 61 evaluated quality models; Table~\ref{tab:revisionmodels} separately lists the nine revision additions. Training hours include all update-time measurements along a full recipe; ``new'' hours exclude a reused prefix. Allocated memory is the maximum over prefix and suffix for inherited runs, rather than only the smaller post-switch value. These are not driver-memory or energy measurements. The full-recipe column is appropriate for recipe comparisons; summing it would double-count shared prefixes.

The successive P1, P2, and P3 extension stages processed 35,075,391,488 new target tokens in 277.652 recorded training GPU-hours. This subtotal excludes the earlier base/control stages and engineering gates. The final 156-job evaluation campaign accounts for 77.544 GPU-process hours, which include loading, hashing, data handling, and evaluation, but exclude separately recorded engineering prechecks and timing probes. These numbers have different accounting scopes and are not presented as a single total project compute estimate.

\section{Complete Depth Curves}
\label{app:curves}
Figure~\ref{fig:depth} and Tables~\ref{tab:curves1}--\ref{tab:curves2} include all 366 endpoint depth measurements. These curves use the development pool. They were not used to select test depths; all locked-test and downstream results remain fixed at \Te$=16$. The original raw endpoint tables retain all 47 non-switching checkpoints, including unfavorable controls, with no winner-only filtering. The prior nine-model revision and the 36-model strengthening extension have no added six-depth curves and must not be counted in these 366 measurements.

\section{Detailed Downstream and Window Comparisons}
\label{app:downstreamdetail}

\begin{table*}[t]
\centering\papertable{\begin{tabular}{@{}lllrr@{}}
\toprule
Cohort & Budget & Training seeds & $\Delta$ LAMBADA (pp) [95\%] & $\Delta$ CORE $\times100$ [95\%] \\
\midrule
Exploratory & 1x & 1102,1103,1104 & +0.375 [-0.149, +0.906] & +0.256 [-0.276, +0.786] \\
Replication & 1x & 1105,1106 & -0.670 [-1.291, -0.039] & +0.935 [+0.218, +1.642] \\
Exploratory & 4x & 1102,1104 & +1.038 [+0.369, +1.698] & +2.246 [+1.478, +3.019] \\
Replication & 4x & 1105,1106 & +0.699 [+0.010, +1.398] & +0.254 [-0.356, +0.866] \\
\bottomrule
\end{tabular}}
\caption{\textbf{Downstream preferences vary across budgets and training-seed cohorts.} Each difference is K8 minus K4+ at approximately matched training time, averaged over the listed seeds. Positive values favor K8. CORE differences are in chance-adjusted score points after multiplying by 100; LAMBADA differences are in final-word accuracy percentage points. The $1\times$ and $4\times$ budgets use the update pairs in Figure~\ref{fig:persistence}. Brackets give marginal 95\% paired source-bootstrap intervals conditional on the evaluated checkpoints, not intervals over future training seeds. Individual pairs appear in Table~\ref{tab:revisionpairs}; seed standard deviations appear in Table~\ref{tab:seedsummary}.}
\label{tab:cohorts}
\end{table*}\label{app:revision}

\begin{figure*}[t]
\centering\includegraphics[width=\textwidth]{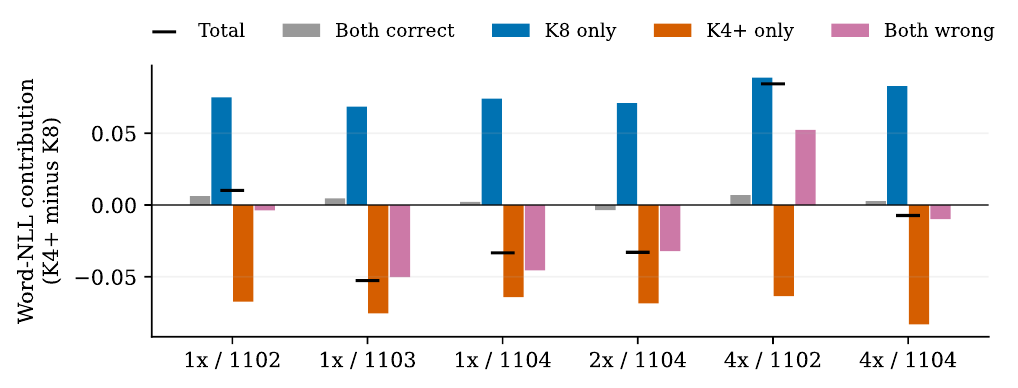}
\caption{Word-NLL contributions from all four correctness groups in all six exploratory pairs. Each sum is divided by all 5,153 examples; signed bars reconstruct the total (black mark). Negative favors K4+. Group membership is checkpoint-dependent, and these are descriptive output decompositions, not causal mechanisms.}
\label{fig:contributions}
\end{figure*}

The complete common-three-seed matched-time comparison appears in Table~\ref{tab:windows} in the main text. Its paired source intervals remain in Table~\ref{tab:windowpairs}.

\paragraph{LAMBADA changes direction at the short budget.}
The exploratory mean favors K8 by 0.375 pp, whereas both replication seeds favor K4+, with a mean difference of $-0.669$ pp. At the long budget all four point estimates favor K8, but three individual source intervals cross zero. The replication-cohort two-seed mean is $+0.699$ pp, with a lower bound of only $+0.010$ pp. Its narrow exclusion of zero is checkpoint-conditional, not a demonstrated population-level training effect.

Compare budgets within a seed: let $D$ be K8 minus K4+ accuracy and $J=D_{\mathrm{long}}-D_{\mathrm{short}}$. For replication seeds 1105/1106, $J$ is $+1.494$ [$+0.233,+2.775$] and $+1.242$ [$-0.039,+2.523$] pp. Both observed changes are positive, but one interval crosses zero. Common source weights preserve pairing across budgets. Subtracting unrelated cohort means would answer a different question.

\paragraph{The large long-budget CORE effect attenuates.}
The exploratory long-budget mean is $+2.246$ CORE$\times100$, compared with $+0.254$ in the replication seeds. Their individual differences are $-0.148$ and $+0.657$; both source intervals include zero. The large original effect does not consistently replicate. The replication-cohort short-budget mean is $+0.935$, and both replication-cohort CORE budget changes are negative with intervals spanning zero (Table~\ref{tab:budgetchanges}). CORE therefore does not follow the LAMBADA budget pattern.

All five CORE family point differences favored K8 in each exploratory long-budget seed. Replication-cohort families are mixed: reading-comprehension differences are $+0.648$ and $-1.060$ in family score units. Complete task rows, families, and crop decompositions are retained in Appendix~\ref{app:eval}. No favorable task is selected to rescue the broad claim. Pooling the two cohorts gives short-budget LAMBADA near zero ($-0.043$ pp) and long-budget CORE $+1.250$; these descriptive means must not conceal the cohort differences.

\subsection{Joint word-loss and accuracy evidence}
Four of six exploratory pairs have lower LAMBADA word NLL under K4+ but higher accuracy under K8. None of their approximately 95\% joint rectangles lies wholly in that opposing-preference quadrant (Figure~\ref{fig:joint}). The one jointly resolved comparison, long-budget seed 1102, favors K8 on both metrics: word-NLL difference $+0.08445$ [$+0.04320,+0.12476$] and accuracy difference $+1.649$ [$+0.582,+2.698$] pp. Its exact McNemar $p=0.000563$ becomes 0.003377 after six-pair Holm adjustment.

The other accuracy intervals cross zero. This neither proves equivalence nor rules out true opposing rankings; the earlier point-estimate disagreements are simply not jointly resolved enough to support a general trade-off. Per-pair rectangles also do not confer simultaneous six-pair coverage.

\subsection{Offsetting correctness and loss contributions}

Short-budget seed 1103 has 280 examples correct only for K8 and 277 correct only for K4+: 557 discordant examples produce a net difference of three. Its exact McNemar $p=0.9325$, rather than evidence of a resolved advantage. The four word-NLL contributions are $+0.004589$ (both correct), $+0.068533$ (K8 only), $-0.075481$ (K4+ only), and $-0.050293$ (both wrong). The single-correct groups largely cancel; jointly incorrect examples account for most of the net K4+ word-loss gain.

Several other pairs show a negative both-wrong contribution, but long-budget seed 1102 reverses it ($+0.052338$, favoring K8). Tokenization strata also differ: in short-budget 1103, multi-token targets contribute $-0.054459$ nats/word and single-token targets $+0.001808$, with opposing accuracy contributions. These identify where measured differences arise, not a universal tokenization effect. Teacher-forced token margins are reported in Appendix~\ref{app:uncertainty}; no calibration experiment identifies overconfidence or a neural mechanism.

\subsection{K4/K6/K8 at the same short-time budget}
\label{app:windowdetail}

In seed 1102, K6 has higher CORE and LAMBADA than both alternatives but higher test NLL than K4+. Its favorable task ordering does not recur consistently (Table~\ref{tab:windows}). In 1104, K4+ has lower test loss and higher CORE and LAMBADA than K6, but is slightly slower. All three common-three-seed paired task-score intervals cross zero (Table~\ref{tab:windowpairs}). Tiny timing differences prevent strict four-objective dominance even when the quality point estimates favor one recipe. The finite set therefore supplies no consistent multi-metric winner.

\begin{figure*}[t]
\centering\includegraphics[width=\textwidth]{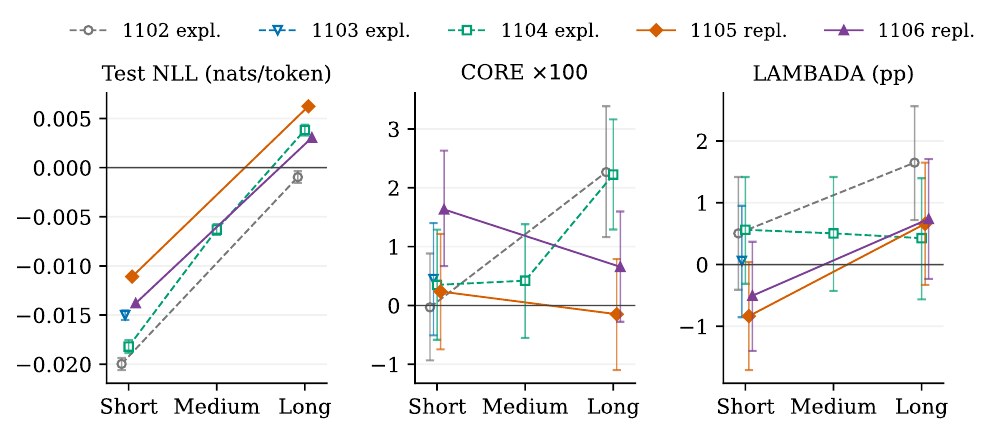}
\caption{\textbf{Budget comparisons retain individual exploratory and replication seeds.} Positive favors K8: test NLL is K4+ minus K8; task scores use K8 minus K4+. Lines connect separately trained endpoints at the measured budgets within the same seed, not a fitted crossover. Downstream bars are marginal 95\% source-cluster intervals. Exploratory NLL bars use paired test blocks; replication NLL points retain the historical display. Updated paired-block intervals appear in Appendix~\ref{app:gr-controls}. Dashed/open: exploratory; solid/filled: replication.}
\label{fig:downstreammetrics}
\end{figure*}
\begin{figure*}[t]
\centering\includegraphics[width=\textwidth]{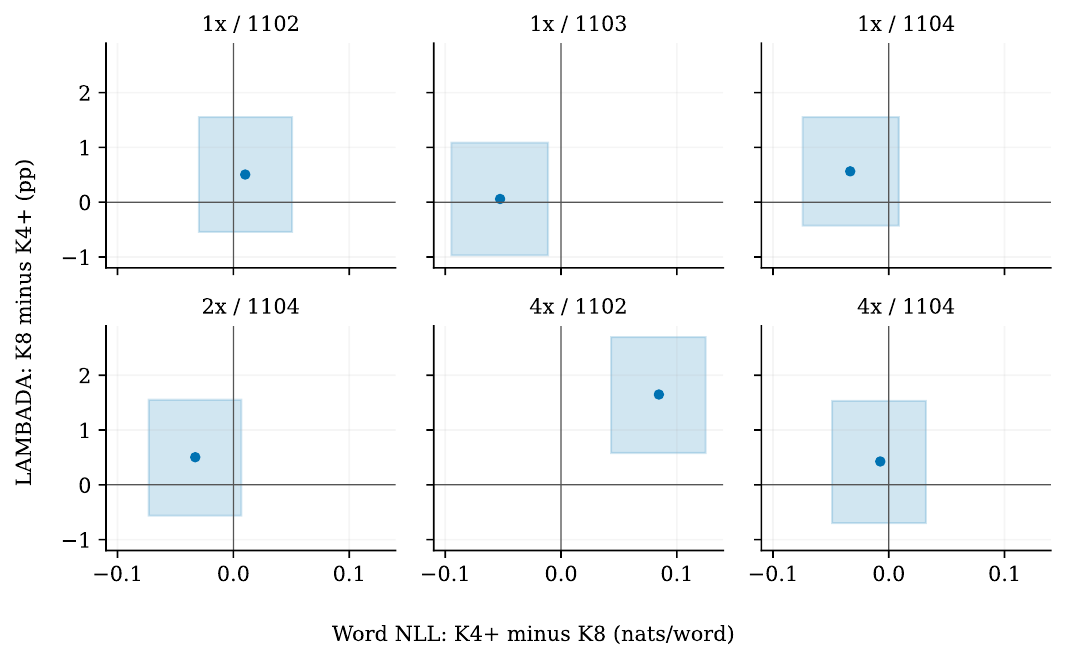}
\caption{All six exploratory LAMBADA pairs and approximate 95\% per-pair joint rectangles (two 97.5\% marginal source-cluster intervals). The upper-left quadrant is the proposed loss-K4+/accuracy-K8 opposition; no rectangle is wholly contained there. Only 4x/1102 is jointly resolved, favoring K8 on both axes. These are not simultaneous six-pair bounds.}
\label{fig:joint}
\end{figure*}

\clearpage
\begin{table*}[p]
\centering\papertable{\begin{tabular}{@{}llrrrr@{}}
\toprule
Cohort & Budget/seed & $\Delta$ test NLL & Time ratio & $\Delta$ LAMBADA (pp) [95\%] & $\Delta$ CORE $\times100$ [95\%] \\
\midrule
Expl. & 1x/1102 & -0.01997003 & 1.002771 & +0.505 [-0.408, +1.417] & -0.032 [-0.933, +0.883] \\
Expl. & 1x/1103 & -0.01500560 & 1.001131 & +0.058 [-0.854, +0.951] & +0.449 [-0.507, +1.401] \\
Expl. & 1x/1104 & -0.01820345 & 1.000533 & +0.563 [-0.310, +1.417] & +0.350 [-0.583, +1.289] \\
Expl. & 2x/1104 & -0.00629726 & 1.000202 & +0.505 [-0.427, +1.417] & +0.421 [-0.552, +1.380] \\
Expl. & 4x/1102 & -0.00096407 & 1.001499 & +1.650 [+0.718, +2.562] & +2.267 [+1.162, +3.386] \\
Expl. & 4x/1104 & +0.00382611 & 0.999041 & +0.427 [-0.563, +1.397] & +2.224 [+1.292, +3.162] \\
Replication & 1x/1105 & -0.01109037 & 0.998533 & -0.834 [-1.708, +0.039] & +0.238 [-0.746, +1.217] \\
Replication & 4x/1105 & +0.00624561 & 1.007399 & +0.660 [-0.330, +1.650] & -0.148 [-1.098, +0.787] \\
Replication & 1x/1106 & -0.01379741 & 1.004858 & -0.505 [-1.398, +0.369] & +1.632 [+0.670, +2.632] \\
Replication & 4x/1106 & +0.00305273 & 1.003046 & +0.737 [-0.233, +1.708] & +0.657 [-0.277, +1.597] \\
\bottomrule
\end{tabular}}
\caption{All ten approximately matched-time comparisons. Positive differences favor K8; time is K4+/K8. Downstream intervals are marginal 95\% source-cluster intervals. The historical point estimates are unchanged; updated paired test-block intervals appear in Appendix~\ref{app:gr-controls}.}
\label{tab:revisionpairs}
\end{table*}
\begin{table*}[p]
\centering\papertable{\begin{tabular}{@{}llrrr@{}}
\toprule
Cohort/budget & Seeds & $\Delta$ word NLL; SD & $\Delta$ LAMBADA; SD & $\Delta$ CORE$\times100$; SD \\
\midrule
Expl./1x & 1102,1103,1104 & -0.02524; 0.03220 & +0.375; 0.276 & +0.256; 0.254 \\
Expl./4x & 1102,1104 & +0.03857; 0.06489 & +1.038; 0.865 & +2.246; 0.030 \\
Repl/1x & 1105,1106 & -0.06745; 0.07910 & -0.670; 0.233 & +0.935; 0.986 \\
Repl/4x & 1105,1106 & +0.06037; 0.01975 & +0.699; 0.055 & +0.254; 0.569 \\
Pooled/1x & 1102,1103,1104,1105,1106 & -0.04212; 0.05116 & -0.043; 0.616 & +0.527; 0.643 \\
Pooled/4x & 1102,1104,1105,1106 & +0.04947; 0.04113 & +0.868; 0.537 & +1.250; 1.196 \\
\bottomrule
\end{tabular}}
\caption{Fixed-cohort point differences and sample SD across training seeds, listed separately from source-sample intervals in Table~\ref{tab:cohorts}. Pooled rows are descriptive. LAMBADA is in pp; CORE is multiplied by 100; word loss is nats/word.}
\label{tab:seedsummary}
\end{table*}
\clearpage
\begin{table*}[p]
\centering\papertable{\begin{tabular}{@{}llr@{}}
\toprule
Seed & Metric & $J=D_{\mathrm{long}}-D_{\mathrm{short}}$ [95\%] \\
\midrule
1105 & LAMBADA (pp) & +1.494 [+0.233, +2.775] \\
1105 & CORE $\times100$ & -0.386 [-1.743, +0.961] \\
1106 & LAMBADA (pp) & +1.242 [-0.039, +2.523] \\
1106 & CORE $\times100$ & -0.975 [-2.379, +0.402] \\
\bottomrule
\end{tabular}
}
\caption{Within-replication-cohort downstream budget changes using common source weights across short and long conditions. Positive means a greater K8-minus-K4+ task difference at long budget. These are complete-recipe contrasts, not token-only causal effects.}
\label{tab:budgetchanges}
\end{table*}
\begin{table*}[p]
\centering\papertable{\begin{tabular}{@{}lrrr@{}}
\toprule
A / B & Test NLL A$-$B & LAMBADA B$-$A (pp) [95\%] & CORE B$-$A $\times100$ [95\%] \\
\midrule
K4/4986 / K6/4498 & -0.002847 & -0.058 [-0.582, +0.453] & -0.136 [-0.653, +0.383] \\
K6/4498 / K8/4096 & -0.014879 & +0.433 [-0.045, +0.931] & +0.392 [-0.102, +0.892] \\
K4/4986 / K8/4096 & -0.017726 & +0.375 [-0.149, +0.906] & +0.256 [-0.276, +0.786] \\
\bottomrule
\end{tabular}}
\caption{Common seeds 1102/1103/1104 only. All three pairwise task-score intervals cross zero. Test-loss differences are means of full-precision endpoints; no new NLL uncertainty is inferred.}
\label{tab:windowpairs}
\end{table*}
\clearpage
\begin{table*}[p]
\centering\papertable{\begin{tabular}{@{}lrrrrrr@{}}
\toprule
Budget/seed & Both correct & K8 only & K4+ only & Both wrong & Exact $p$ & Holm $p$ \\
\midrule
1x/1102 & 892 & 296 & 270 & 3695 & 0.293330 & 1.000000 \\
1x/1103 & 885 & 280 & 277 & 3711 & 0.932476 & 1.000000 \\
1x/1104 & 933 & 278 & 249 & 3693 & 0.222543 & 1.000000 \\
2x/1104 & 996 & 300 & 274 & 3583 & 0.296720 & 1.000000 \\
4x/1102 & 1090 & 340 & 255 & 3468 & 0.000563 & 0.003377 \\
4x/1104 & 1058 & 336 & 314 & 3445 & 0.410136 & 1.000000 \\
\bottomrule
\end{tabular}}
\caption{Actual LAMBADA correctness-transition counts for all six old pairs. Exact McNemar is an independent-example sensitivity test; Holm correction covers these six comparisons only.}
\label{tab:discordance}
\end{table*}
\begin{table*}[p]
\centering\papertable{\begin{tabular}{@{}lrrrrr@{}}
\toprule
Budget/seed & Both correct & K8 only & K4+ only & Both wrong & Total \\
\midrule
1x/1102 & +0.006159 & +0.075054 & -0.067252 & -0.003738 & +0.010223 \\
1x/1103 & +0.004589 & +0.068533 & -0.075481 & -0.050293 & -0.052651 \\
1x/1104 & +0.002225 & +0.074146 & -0.064207 & -0.045452 & -0.033288 \\
2x/1104 & -0.003457 & +0.071094 & -0.068499 & -0.032060 & -0.032923 \\
4x/1102 & +0.006864 & +0.088740 & -0.063487 & +0.052338 & +0.084454 \\
4x/1104 & +0.002792 & +0.082844 & -0.083181 & -0.009771 & -0.007316 \\
\bottomrule
\end{tabular}}
\caption{Signed word-NLL contributions, K4+ minus K8, each divided by all 5,153 examples. Four groups reconstruct the total; small differences between printed sums reflect rounding.}
\label{tab:contributions}
\end{table*}
\begin{table*}[p]
\centering\papertable{\begin{tabular}{@{}llrrrrr@{}}
\toprule
Seed & Window/updates & Targets & Hours & Test NLL & CORE$\times100$ & LAMBADA\% \\
\midrule
1105 & K8/16384 & 2,147,483,648 & 20.7357 & 3.05671211 & 12.5069 & 27.3627 \\
1106 & K8/16384 & 2,147,483,648 & 20.8029 & 3.05747374 & 12.8954 & 27.2851 \\
1105 & K4/19944 & 2,614,099,968 & 20.8891 & 3.06295772 & 12.6553 & 26.7029 \\
1106 & K4/19944 & 2,614,099,968 & 20.8663 & 3.06052647 & 12.2387 & 26.5476 \\
1105 & K8/4096 & 536,870,912 & 5.2078 & 3.26282599 & 10.0307 & 21.7737 \\
1106 & K8/4096 & 536,870,912 & 5.2007 & 3.26284120 & 10.6667 & 21.5408 \\
1105 & K4/4986 & 653,524,992 & 5.2001 & 3.25173562 & 9.7931 & 22.6082 \\
1106 & K4/4986 & 653,524,992 & 5.2260 & 3.24904379 & 9.0348 & 22.0454 \\
1104 & K6/4498 & 589,561,856 & 5.1944 & 3.25737316 & 10.4328 & 22.4141 \\
\bottomrule
\end{tabular}}
\caption{Nine endpoints from the prior experimental revision, each from one completed run per configuration--seed. All use native mean forward depth 16 and fixed T16 evaluation. These additions do not have six-depth curves; full-precision word loss/perplexity and hashes are in the revision CSV.}
\label{tab:revisionmodels}
\end{table*}
\clearpage
\begin{table*}[p]
\centering\papertable{\begin{tabular}{@{}lrrrrr@{}}
\toprule
Task / family & 1102 & 1103 & 1104 & 1105 & 1106 \\
\midrule
hellaswag\_zeroshot & -0.199 & -0.305 & -0.146 & -0.305 & -0.969 \\
jeopardy & -0.142 & -0.016 & +0.283 & +0.205 & -0.504 \\
bigbench\_qa\_wikidata & +0.130 & +0.377 & -2.483 & -5.822 & -9.212 \\
arc\_easy & +0.991 & +1.141 & -0.299 & +0.262 & +0.842 \\
arc\_challenge & -1.100 & +1.176 & +0.152 & +0.493 & -1.744 \\
copa & +0.000 & +0.000 & +0.000 & -10.000 & +12.000 \\
commonsense\_qa & +7.166 & +8.736 & -7.917 & +2.491 & +6.279 \\
piqa & -2.612 & -2.612 & -0.871 & +2.213 & -1.886 \\
openbook\_qa & -0.533 & -0.267 & +2.667 & +0.000 & +3.200 \\
lambada\_openai & +0.078 & +1.067 & +1.223 & -1.184 & -1.533 \\
hellaswag & -0.292 & -0.186 & +0.181 & -0.124 & -0.531 \\
winograd & -6.593 & -3.663 & +6.593 & +4.396 & +1.465 \\
winogrande & -1.894 & +0.789 & -6.156 & +1.263 & +1.579 \\
bigbench\_dyck\_languages & +1.933 & +0.067 & +0.800 & +0.467 & -1.167 \\
agi\_eval\_lsat\_ar & +0.543 & -0.543 & -0.181 & -1.812 & -2.355 \\
bigbench\_cs\_algorithms & -0.202 & -3.611 & -1.818 & +4.545 & -2.424 \\
bigbench\_operators & -0.794 & -0.159 & +5.238 & -0.476 & -4.921 \\
bigbench\_repeat\_copy\_logic & -1.042 & +2.083 & +0.000 & +2.083 & +4.167 \\
squad & +0.874 & +1.164 & +1.741 & +1.637 & +1.006 \\
coqa & +0.864 & +2.217 & +2.054 & +0.576 & -0.601 \\
boolq & +2.575 & +2.575 & +6.519 & +4.480 & +33.183 \\
bigbench\_language\_identification & -0.455 & -0.150 & +0.125 & -0.161 & +0.029 \\
Family: commonsense\_reasoning & +1.005 & +1.464 & -1.530 & -1.324 & +4.898 \\
Family: language\_understanding & -1.559 & -0.408 & +0.303 & +0.647 & +0.007 \\
Family: reading\_comprehension & +1.438 & +1.985 & +3.438 & +2.231 & +11.196 \\
Family: symbolic\_problem\_solving & +0.088 & -0.433 & +0.808 & +0.962 & -1.340 \\
Family: world\_knowledge & -0.030 & +0.670 & -0.587 & -1.216 & -2.655 \\
CORE total & -0.032 & +0.449 & +0.350 & +0.238 & +1.632 \\
\bottomrule
\end{tabular}}
\caption{All 22 short-budget task differences and five family means, CORE-style chance adjustment multiplied by 100, K8 minus K4+. Exploratory seeds are 1102/1103/1104; new seeds are 1105/1106. Family means are multiplied by task count/22 to obtain aggregate contributions. The CSV retains every marginal interval.}
\label{tab:revisiontasks1}
\end{table*}
\clearpage
\begin{table*}[p]
\centering\papertable{\begin{tabular}{@{}lrrrr@{}}
\toprule
Task / family & 1102 & 1104 & 1105 & 1106 \\
\midrule
hellaswag\_zeroshot & +1.540 & +0.850 & +0.119 & -0.013 \\
jeopardy & +1.228 & +0.661 & -0.047 & -0.236 \\
bigbench\_qa\_wikidata & +2.016 & +4.122 & +1.821 & -1.578 \\
arc\_easy & +3.217 & -0.112 & -0.412 & -0.486 \\
arc\_challenge & +1.214 & +1.138 & +2.086 & +0.493 \\
copa & +8.000 & +12.000 & -8.000 & +12.000 \\
commonsense\_qa & -0.819 & +12.080 & -0.717 & +1.092 \\
piqa & +0.399 & -0.472 & +0.363 & -0.326 \\
openbook\_qa & +2.667 & -4.267 & +1.067 & -4.800 \\
lambada\_openai & +1.417 & +0.582 & -0.776 & +0.912 \\
hellaswag & +0.407 & +0.757 & +0.836 & +0.000 \\
winograd & +2.198 & +3.663 & -3.663 & +4.396 \\
winogrande & +6.946 & +5.209 & +0.474 & -2.526 \\
bigbench\_dyck\_languages & -2.133 & -2.233 & -3.600 & +4.367 \\
agi\_eval\_lsat\_ar & -2.717 & -2.536 & +1.812 & +0.543 \\
bigbench\_cs\_algorithms & +3.384 & +4.192 & +0.682 & +4.343 \\
bigbench\_operators & +3.810 & +3.016 & +1.587 & +0.476 \\
bigbench\_repeat\_copy\_logic & +0.000 & -1.042 & +1.042 & -1.042 \\
squad & -0.009 & +1.652 & +2.854 & -0.779 \\
coqa & +1.628 & +1.077 & +0.752 & +1.541 \\
boolq & +15.022 & +8.960 & -1.663 & -3.943 \\
bigbench\_language\_identification & +0.462 & -0.359 & +0.121 & +0.015 \\
Family: commonsense\_reasoning & +2.562 & +4.836 & -1.822 & +1.991 \\
Family: language\_understanding & +2.162 & +1.784 & -0.481 & +0.464 \\
Family: reading\_comprehension & +5.547 & +3.896 & +0.647 & -1.061 \\
Family: symbolic\_problem\_solving & +0.469 & +0.279 & +0.304 & +1.738 \\
Family: world\_knowledge & +1.919 & +1.452 & +0.862 & -0.452 \\
CORE total & +2.267 & +2.224 & -0.148 & +0.657 \\
\bottomrule
\end{tabular}}
\caption{All long-budget task and family point differences; units and weighting match Table~\ref{tab:revisiontasks1}. Exploratory seeds 1102/1104 appear beside replication seeds 1105/1106; mixed new signs remain visible.}
\label{tab:revisiontasks4}
\end{table*}
\clearpage
\setcounter{totalnumber}{5}
\setcounter{topnumber}{5}
\begin{table*}[p]
\centering\papertable{
\begin{tabular}{@{}llrrrrr@{}}
\toprule
Seed & Window & Updates & Test NLL & CORE$\times100$ & LAMB. (\%) & Time / K8 \\
\midrule
1102 & K4 & 4,096 & 3.28879 & 8.842 & 22.550 & 0.8215 \\
1103 & K4 & 4,096 & 3.29723 & 9.992 & 21.133 & 0.8207 \\
1104 & K4 & 4,096 & 3.29518 & 9.379 & 21.754 & 0.8201 \\
1102 & K8 & 4,096 & 3.26746 & 10.077 & 23.055 & 1.0000 \\
1103 & K8 & 4,096 & 3.27054 & 10.110 & 22.608 & 1.0000 \\
1104 & K8 & 4,096 & 3.27203 & 11.029 & 23.501 & 1.0000 \\
1102 & K4 & 4,986 & 3.24749 & 10.109 & 22.550 & 1.0028 \\
1103 & K4 & 4,986 & 3.25553 & 9.661 & 22.550 & 1.0011 \\
1104 & K4 & 4,986 & 3.25382 & 10.679 & 22.938 & 1.0005 \\
\bottomrule
\end{tabular}}
\caption{Unrounded-source per-seed values underlying the three-recipe means in Table~\ref{tab:allocation}. Time ratios use the same seed's K8/4,096 training time.}
\label{tab:allocationseeds}
\end{table*}
\begin{table*}[p]
\centering\papertable{
\begin{tabular}{rrrrrr}
\toprule
Seed & K6 dev & 4$\to$8 dev & 8$\to$4 dev & $C_{48}/C_6$ & $C_{84}/C_6$ \\
\midrule
1102 & 3.26142 & 3.26301 & 3.26626 & 0.9992 & 1.0032 \\
1103 & 3.26344 & 3.27327 & 3.27075 & 0.9987 & 1.0017 \\
1104 & 3.26538 & 3.26983 & 3.27360 & 0.9967 & 0.9979 \\
\bottomrule
\end{tabular}
}
\caption{Constant K6 versus half-run switches at 4,096 total updates: dev NLL and realized switch/K6 training-time ratios. Complete prefix cost is included for inherited trials. Lower loss favors K6 in every row.}
\label{tab:k6timing}
\end{table*}
\begin{table*}[p]
\centering\papertable{
\begin{tabular}{@{}llrrll@{}}
\toprule
Budget & Seed & $\Delta L$ & $\Delta$ correct & 95\% block CI & LOTO CORE$\times100$ \\
\midrule
1x & 1102 & -0.01997 & +26 & [-0.02060, -0.01937] & [-0.375, +0.281] \\
1x & 1103 & -0.01501 & +3 & [-0.01550, -0.01452] & [+0.055, +0.645] \\
1x & 1104 & -0.01820 & +29 & [-0.01886, -0.01755] & [+0.053, +0.744] \\
2x & 1104 & -0.00630 & +26 & [-0.00684, -0.00576] & [+0.197, +0.918] \\
4x & 1102 & -0.00096 & +85 & [-0.00156, -0.00036] & [+1.660, +2.504] \\
4x & 1104 & +0.00383 & +22 & [+0.00326, +0.00437] & [+1.755, +2.534] \\
\bottomrule
\end{tabular}}
\caption{Post hoc main-pair diagnostics. $\Delta L=L_{4+}-L_8$; positive net-correct counts favor K8 on LAMBADA. NLL intervals use paired contiguous 131,072-target blocks. LOTO is the range of K8 minus extended-K4 CORE$\times100$ scores when each task is omitted in turn; it is not a confidence interval.}
\label{tab:uncertainty}
\end{table*}
\begin{table*}[p]
\centering
\papertable{
\begin{tabular}{rrrrr}
\toprule
Seed & FULL 4096 & K4 4096 & K4 6494 & Cost ratio \\
\midrule
1102 & 3.27467 & 3.28527 & 3.19486 & 0.9915 \\
1103 & 3.27632 & 3.28974 & 3.19909 & 0.9912 \\
\bottomrule
\end{tabular}
}
\caption{Native $\mu=8$: locked-test NLL and extended-\K{4}/FULL actual training-time ratios. Evaluation recurrence is 16 for every entry.}
\label{tab:mu8}
\end{table*}
\clearpage
\begin{table*}[p]
\centering\papertable{
\begin{tabular}{@{}llrrrrr@{}}
\toprule
$\mu$ & Budget/seed & $\Delta$ test NLL & $\Delta$ CORE$\times100$ & $\Delta$ LAMB. (pp) & PPL K4+ & PPL ref. \\
\midrule
16 & 1x/1102 & -0.01997 & -0.032 & +0.505 & 140.358 & 138.930 \\
16 & 1x/1103 & -0.01501 & +0.449 & +0.058 & 147.361 & 155.328 \\
16 & 1x/1104 & -0.01820 & +0.350 & +0.563 & 145.522 & 150.448 \\
16 & 2x/1104 & -0.00630 & +0.421 & +0.505 & 98.249 & 101.538 \\
16 & 4x/1102 & -0.00096 & +2.267 & +1.650 & 80.640 & 74.110 \\
16 & 4x/1104 & +0.00383 & +2.224 & +0.427 & 75.092 & 75.644 \\
8 & FULL/1102 & -0.07981 & -0.618 & +0.349 & 119.885 & 136.094 \\
8 & FULL/1103 & -0.07723 & -0.066 & -1.650 & 126.773 & 156.918 \\
\bottomrule
\end{tabular}}
\caption{Exploratory matched-time differences: NLL is extended K4 minus the reference; task scores are reference minus extended K4. The reference is K8 at $\mu=16$ and FULL at $\mu=8$. Positive differences favor the reference. CORE is multiplied by 100; accuracy differences are percentage points. Word perplexities are separate endpoint values.}
\label{tab:downstream}
\end{table*}
\begin{table*}[p]
\centering\papertable{
\begin{tabular}{rrrr}
\toprule
$k$ & $N_k$ & Test 1102 & Test 1103 \\
\midrule
1 & 5,910 & 3.28464 & 3.28106 \\
2 & 5,564 & 3.27192 & 3.27098 \\
4 & 4,986 & 3.24749 & 3.25553 \\
6 & 4,498 & 3.25271 & 3.25531 \\
8 & 4,096 & 3.26746 & 3.27054 \\
\bottomrule
\end{tabular}
}
\caption{The finite short-cost frontier. Horizons were selected from controlled timing. Small K4--K6 differences do not identify a universal optimum.}
\label{tab:grid}
\end{table*}
\begin{figure*}[p]
\centering\includegraphics[width=.85\textwidth]{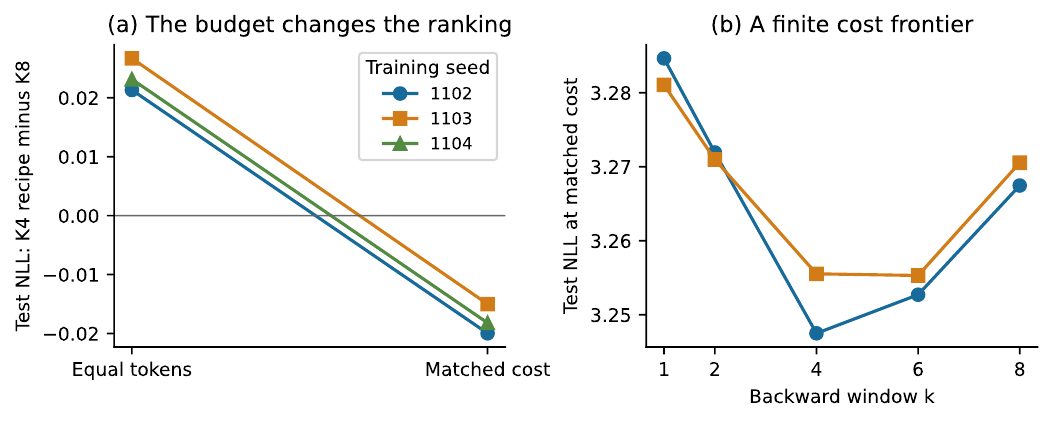}
\caption{Supplementary loss comparisons: the token/time ranking reversal (left) and finite window grid (right). Lines connect separately trained endpoints. All values are locked-test NLL.}
\label{fig:tradeoff}
\end{figure*}
\clearpage
\begin{figure*}[p]
\centering\includegraphics[width=.65\textwidth]{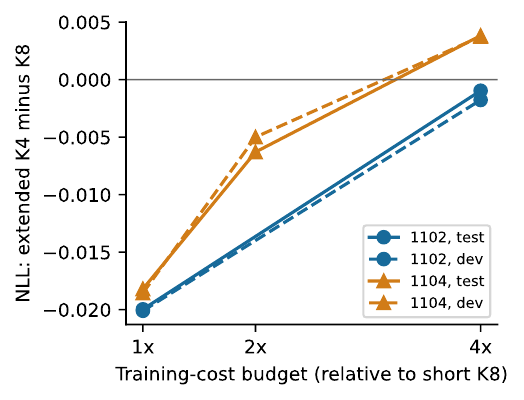}
\caption{Dev and test budget comparisons, extended K4 minus K8. The intermediate budget exists only for seed 1104. Negative favors extended K4.}
\label{fig:long}
\end{figure*}

\begin{table*}[p]
\centering
\papertable{
\begin{tabular}{rrrrrrr}
\toprule
Seed & $L_{44}$ & $L_{88}$ & $L_{84}$ & $L_{48}$ & $R_{\downarrow}$ & $R_{\uparrow}$ \\
\midrule
1102 & 3.275621 & 3.254366 & 3.266256 & 3.263006 & 0.4406 & 0.5935 \\
1103 & 3.285229 & 3.259002 & 3.270754 & 3.273271 & 0.5519 & 0.4559 \\
1104 & 3.283440 & 3.261048 & 3.273596 & 3.269833 & 0.4396 & 0.6077 \\
\bottomrule
\end{tabular}}
\caption{Half-run timing endpoints (dev NLL) and raw gap-retention ratios. All switches occur at update 2,048 of 4,096.}
\label{tab:half}
\end{table*}
\begin{table*}[p]
\centering
\papertable{
\begin{tabular}{rrrrrr}
\toprule
Seed & $G$ & $\sum_q g_q-G$ & $g_1+g_2-g_{84}$ & $g_3+g_4-g_{48}$ & $\sum_q w_q$ \\
\midrule
1102 & 0.021255 & +0.000823 & -0.000641 & +0.000739 & 1.038725 \\
1103 & 0.026227 & +0.000075 & -0.000483 & +0.000352 & 1.002842 \\
\bottomrule
\end{tabular}
}
\caption{Raw dev residuals for the three frozen quarter additivity checks. All six absolute residuals are below the per-seed $0.25G$ threshold.}
\label{tab:timing}
\end{table*}

\clearpage
\begin{table*}[p]
\centering\small
\papertable{
\begin{tabular}{lrrr}
\toprule
Task & Examples & Shots & $b_t$ \\
\midrule
hellaswag\_zeroshot & 10,042 & 0 & 0.2500 \\
jeopardy & 2,117 & 10 & 0.0000 \\
bigbench\_qa\_wikidata & 20,321 & 10 & 0.0000 \\
arc\_easy & 2,376 & 10 & 0.2500 \\
arc\_challenge & 1,172 & 10 & 0.2500 \\
copa & 100 & 0 & 0.5000 \\
commonsense\_qa & 1,221 & 10 & 0.2000 \\
piqa & 1,838 & 10 & 0.5000 \\
openbook\_qa & 500 & 0 & 0.2500 \\
lambada\_openai & 5,153 & 0 & 0.0000 \\
hellaswag & 10,042 & 10 & 0.2500 \\
winograd & 273 & 0 & 0.5000 \\
winogrande & 1,267 & 0 & 0.5000 \\
bigbench\_dyck\_languages & 1,000 & 10 & 0.0000 \\
agi\_eval\_lsat\_ar & 230 & 3 & 0.2000 \\
bigbench\_cs\_algorithms & 1,320 & 10 & 0.0000 \\
bigbench\_operators & 210 & 10 & 0.0000 \\
bigbench\_repeat\_copy\_logic & 32 & 10 & 0.0000 \\
squad & 10,570 & 10 & 0.0000 \\
coqa & 7,983 & 0 & 0.0000 \\
boolq & 3,270 & 10 & 0.6200 \\
bigbench\_language\_identification & 10,000 & 10 & 0.0910 \\
\bottomrule
\end{tabular}}
\caption{Complete CORE (adapted) task inventory and fixed chance baselines. Counts are per prompt seed; identical zero-shot predictions are reused. The accompanying frozen configuration specifies the task-specific shot counts and formats.}
\label{tab:tasks}
\end{table*}
\begin{figure*}[p]
\centering\includegraphics[width=\textwidth]{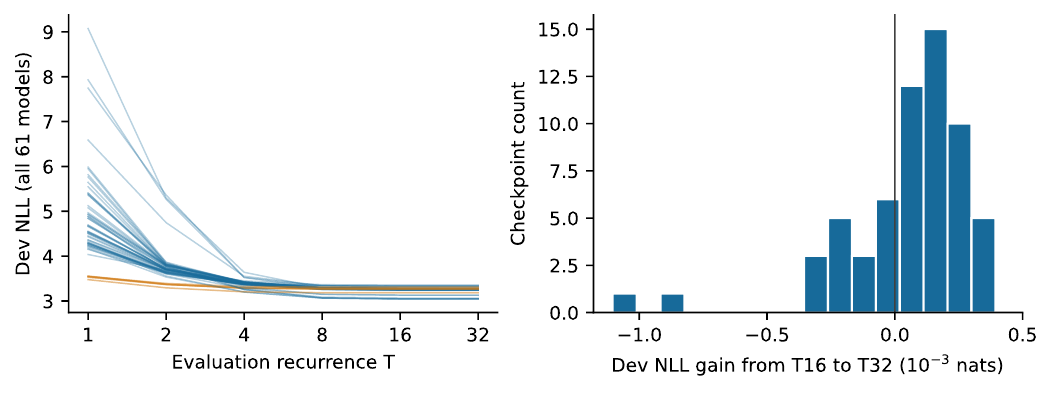}
\caption{Inference-depth coverage for the original 61 final models. Left: each line is one final model (blue: native $\mu=16$; orange: native $\mu=8$). Right: raw marginal dev-NLL gain from recurrence 16 to 32. The narrow terminal range does not establish a universal ceiling.}
\label{fig:depth}
\end{figure*}
\clearpage
\begin{table*}[p]
\centering\small
\papertable{
\begin{tabular}{llrrrrrrr}
\toprule
ID & Recipe & $\mu$ & Seed & $N$ & Tokens (M) & Train (h) & New (h) & Peak (GiB) \\
\midrule
M01 & FULL & 16 & 1102 & 4,096 & 536.871 & 8.420 & 8.420 & 41.01 \\
M02 & FULL & 16 & 1103 & 4,096 & 536.871 & 8.377 & 8.377 & 39.21 \\
M03 & K4 & 16 & 1102 & 4,096 & 536.871 & 4.269 & 4.269 & 10.47 \\
M04 & K4 & 16 & 1103 & 4,096 & 536.871 & 4.273 & 4.273 & 10.47 \\
M05 & MLMC & 16 & 1102 & 4,096 & 536.871 & 4.831 & 4.831 & 41.01 \\
M06 & MLMC & 16 & 1103 & 4,096 & 536.871 & 4.827 & 4.827 & 38.32 \\
M07 & K6 & 16 & 1102 & 4,096 & 536.871 & 4.744 & 4.744 & 12.27 \\
\bottomrule
\end{tabular}}
\caption{Original evaluated-model ledger, part 1. Q$q$ denotes \K{8} in quarter $q$ and \K{4} elsewhere. Half-run recipes switch at the midpoint. Full recipe token count includes reused prefixes. Peak is full-recipe allocated memory. IDs map to source hashes and original run identifiers in the accompanying evidence.}
\label{tab:ledger1}
\end{table*}
\clearpage
\begin{table*}[p]
\centering\small
\papertable{
\begin{tabular}{llrrrrrrr}
\toprule
ID & Recipe & $\mu$ & Seed & $N$ & Tokens (M) & Train (h) & New (h) & Peak (GiB) \\
\midrule
M32 & 4-to-8 & 16 & 1104 & 4,096 & 536.871 & 4.733 & 2.598 & 14.07 \\
M33 & K4 & 16 & 1104 & 4,986 & 653.525 & 5.207 & 5.207 & 10.47 \\
M34 & 8-to-4 & 16 & 1104 & 4,096 & 536.871 & 4.739 & 2.136 & 14.07 \\
M35 & K6 & 16 & 1102 & 4,498 & 589.562 & 5.206 & 5.206 & 12.27 \\
M36 & K6 & 16 & 1103 & 4,498 & 589.562 & 5.197 & 5.197 & 12.27 \\
M37 & K1 & 16 & 1102 & 5,910 & 774.636 & 5.208 & 5.208 & 7.78 \\
M38 & K1 & 16 & 1103 & 5,910 & 774.636 & 5.212 & 5.212 & 7.78 \\
\bottomrule
\end{tabular}}
\caption{Original evaluated-model ledger, part 2; conventions match Table~\ref{tab:ledger1}. All listed quality arms completed with zero skipped updates.}
\label{tab:ledger2}
\end{table*}
\clearpage
\begin{table*}[p]
\centering\small
\papertable{
\begin{tabular}{@{}lrrrrr@{}}
\toprule
ID & Dev NLL & Test NLL & CORE$\times100$ & LAMB. (\%) & Word PPL \\
\midrule
M01 & 3.267488 & 3.279400 & 9.3415 & 22.5888 & 157.468 \\
M02 & 3.255099 & 3.267241 & 10.9245 & 22.8023 & 152.661 \\
M03 & 3.275621 & 3.288787 & 8.8420 & 22.5500 & 153.365 \\
M04 & 3.285229 & 3.297230 & 9.9916 & 21.1333 & 173.125 \\
M05 & 3.354463 & 3.367380 & 9.2386 & 20.3765 & 223.646 \\
M06 & 3.322300 & 3.334626 & 8.9851 & 19.6002 & 235.979 \\
M07 & 3.261419 & 3.274027 & 10.4612 & 22.4141 & 158.195 \\
\bottomrule
\end{tabular}}
\caption{Unfiltered final T16 scores for the original 47 non-switching models. CORE is the adapted score multiplied by 100; independent LAMBADA reports exact final-word accuracy and word perplexity. NLL is in nats per target token. IDs are defined in the ledger. The 14 temporal ablations have no downstream or test scores by the frozen inclusion rule.}
\label{tab:raw1}
\end{table*}
\clearpage
\begin{table*}[p]
\centering\small
\papertable{
\begin{tabular}{lrrrrrr}
\toprule
ID & T1 & T2 & T4 & T8 & T16 & T32 \\
\midrule
M01 & 4.219654 & 3.650951 & 3.381724 & 3.281824 & 3.267488 & 3.268355 \\
M02 & 4.289251 & 3.657965 & 3.376061 & 3.270260 & 3.255099 & 3.256204 \\
M03 & 5.763369 & 3.836362 & 3.428385 & 3.294597 & 3.275621 & 3.275373 \\
M04 & 4.659656 & 3.724482 & 3.403194 & 3.300035 & 3.285229 & 3.285103 \\
M05 & 4.361200 & 3.701965 & 3.450244 & 3.364940 & 3.354463 & 3.354627 \\
M06 & 4.250415 & 3.658467 & 3.417975 & 3.333620 & 3.322300 & 3.322614 \\
M07 & 4.452059 & 3.696118 & 3.388361 & 3.277871 & 3.261419 & 3.261269 \\
\bottomrule
\end{tabular}}
\caption{Dev depth-curve values for the original 61 models, part 1. Recurrence counts are forward inference depths, not backward windows. All T16 entries exactly reproduce the archived training endpoints before rounding.}
\label{tab:curves1}
\end{table*}
\clearpage
\begin{table*}[p]
\centering\small
\papertable{
\begin{tabular}{lrrrrrr}
\toprule
ID & T1 & T2 & T4 & T8 & T16 & T32 \\
\midrule
M32 & 4.274098 & 3.635656 & 3.377307 & 3.283925 & 3.269833 & 3.269731 \\
M33 & 5.645128 & 3.811015 & 3.399295 & 3.262036 & 3.242494 & 3.242100 \\
M34 & 5.363500 & 3.824905 & 3.428297 & 3.293679 & 3.273596 & 3.273405 \\
M35 & 4.695611 & 3.717222 & 3.375783 & 3.258515 & 3.241241 & 3.240967 \\
M36 & 4.546340 & 3.695171 & 3.371692 & 3.261254 & 3.244283 & 3.244054 \\
M37 & 4.535534 & 3.704598 & 3.350763 & 3.278201 & 3.273075 & 3.272999 \\
M38 & 4.554107 & 3.706297 & 3.342862 & 3.274216 & 3.269355 & 3.269275 \\
\bottomrule
\end{tabular}}
\caption{Dev depth-curve values for the original 61 models, part 2; conventions match Table~\ref{tab:curves1}.}
\label{tab:curves2}
\end{table*}

\clearpage
\begin{table*}[p]
\centering
\papertable{
\begin{tabular}{@{}lrrrrrr@{}}
\toprule
Task / family & 1x/1102 & 1x/1103 & 1x/1104 & 2x/1104 & 4x/1102 & 4x/1104 \\
\midrule
hellaswag\_zeroshot & -0.199 & -0.305 & -0.146 & -0.757 & +1.540 & +0.850 \\
jeopardy & -0.142 & -0.016 & +0.283 & +0.583 & +1.228 & +0.661 \\
bigbench\_qa\_wikidata & +0.130 & +0.377 & -2.483 & +1.376 & +2.016 & +4.122 \\
arc\_easy & +0.991 & +1.141 & -0.299 & +1.684 & +3.217 & -0.112 \\
arc\_challenge & -1.100 & +1.176 & +0.152 & +0.910 & +1.214 & +1.138 \\
copa & -0.000 & -0.000 & -0.000 & -10.000 & +8.000 & +12.000 \\
commonsense\_qa & +7.166 & +8.736 & -7.917 & +2.321 & -0.819 & +12.080 \\
\bottomrule
\end{tabular}}
\caption{Complete task-level and family-level differences for native $\mu=16$: \K{8} minus extended \K{4}, multiplied by 100. Each cell is the full-precision difference averaged over prompt seeds, then rounded for display. Positive favors \K{8}. Task rows use their individual chance-adjusted scores; family rows are within-family means. Family membership is fixed by the original configuration; task order matches Table~\ref{tab:tasks}. The final row is the CORE (adapted) difference multiplied by 100, reconstructed exactly from task scores.}
\label{tab:taskdeltas}
\end{table*}
\clearpage
\begin{table*}[p]
\centering
\papertable{
\begin{tabular}{@{}lrr@{}}
\toprule
Task / family & 1102 & 1103 \\
\midrule
hellaswag\_zeroshot & -1.407 & -0.717 \\
jeopardy & -0.268 & -0.787 \\
bigbench\_qa\_wikidata & +2.575 & -1.575 \\
arc\_easy & -2.488 & -1.253 \\
arc\_challenge & +2.313 & -0.683 \\
copa & -4.000 & -4.000 \\
commonsense\_qa & -2.491 & +2.013 \\
piqa & -1.886 & -3.881 \\
openbook\_qa & -1.067 & -0.000 \\
lambada\_openai & -0.291 & -2.096 \\
hellaswag & -0.841 & -0.938 \\
winograd & +0.733 & +3.663 \\
winogrande & -0.631 & +2.526 \\
bigbench\_dyck\_languages & -0.533 & +10.200 \\
agi\_eval\_lsat\_ar & +0.181 & +2.536 \\
bigbench\_cs\_algorithms & +0.833 & +1.641 \\
bigbench\_operators & +2.222 & -4.603 \\
bigbench\_repeat\_copy\_logic & -0.000 & +1.042 \\
squad & -1.205 & -1.577 \\
coqa & -2.455 & -3.720 \\
boolq & -2.817 & +0.805 \\
bigbench\_language\_identification & -0.084 & -0.051 \\
language understanding & -0.420 & +0.398 \\
world knowledge & +0.533 & -1.074 \\
commonsense reasoning & -2.361 & -1.467 \\
symbolic problem solving & +0.541 & +2.163 \\
reading comprehension & -2.159 & -1.497 \\
CORE$\times100$ & -0.618 & -0.066 \\
\bottomrule
\end{tabular}}
\caption{Complete native-$\mu=8$ task differences, FULL/4,096 minus extended \K{4}/6,494, multiplied by 100. Conventions match Table~\ref{tab:taskdeltas}. These two seeds are paired within native $\mu=8$, not with the $\mu=16$ models. Family means preserve mixed task-level directions rather than assigning a single general capability to each recipe.}
\label{tab:mu8taskdeltas}
\end{table*}

\clearpage
\begin{figure*}[p]
\centering\includegraphics[width=\textwidth]{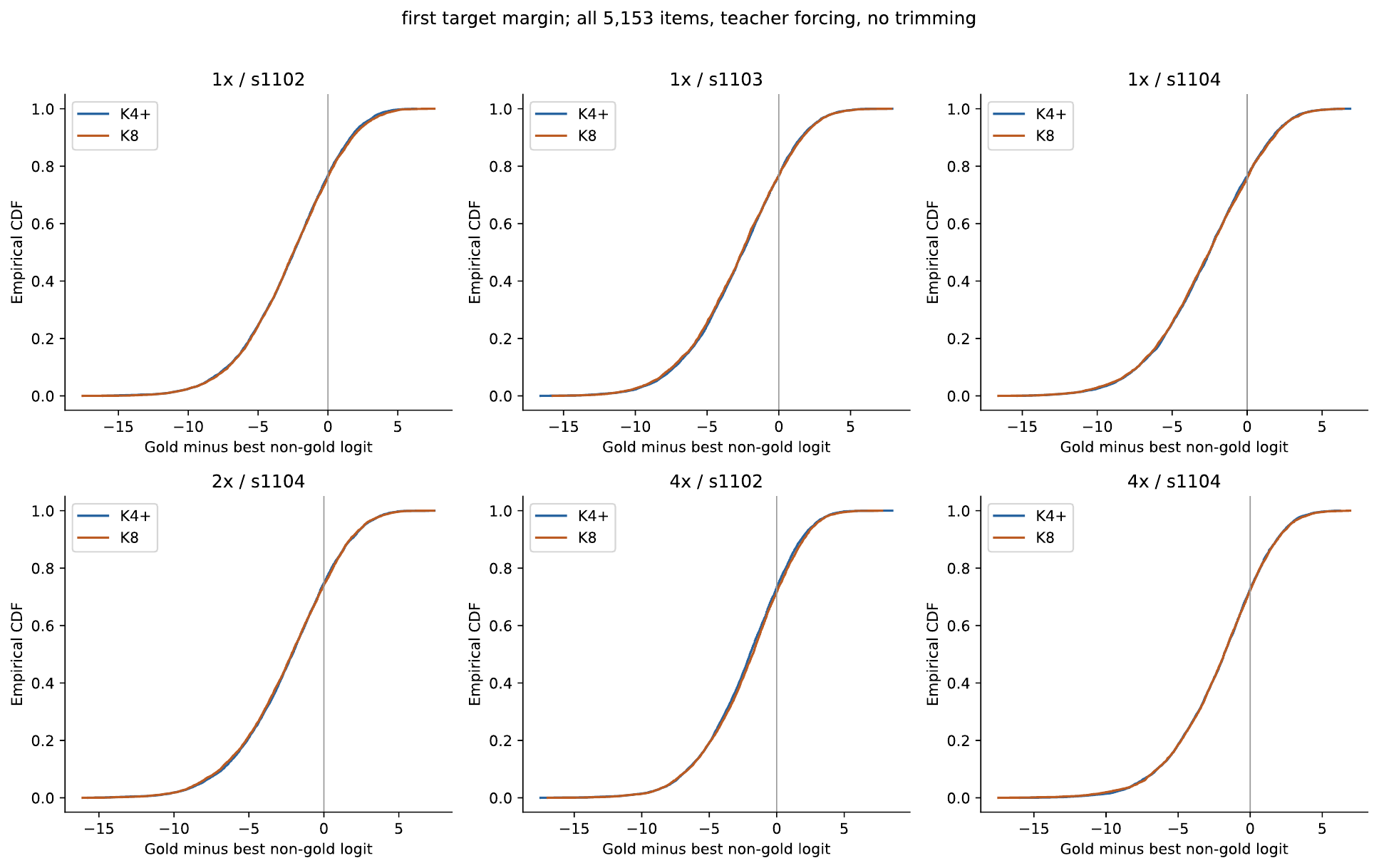}
\caption{First-target-token teacher-forced margin ECDFs for the six exploratory comparisons. Each ECDF uses all 5,153 LAMBADA examples and is shown over the plotted margin range. These distributions do not establish calibration or compare complete word sequences.}
\label{fig:marginfirst}
\end{figure*}
\begin{figure*}[p]
\centering\includegraphics[width=\textwidth]{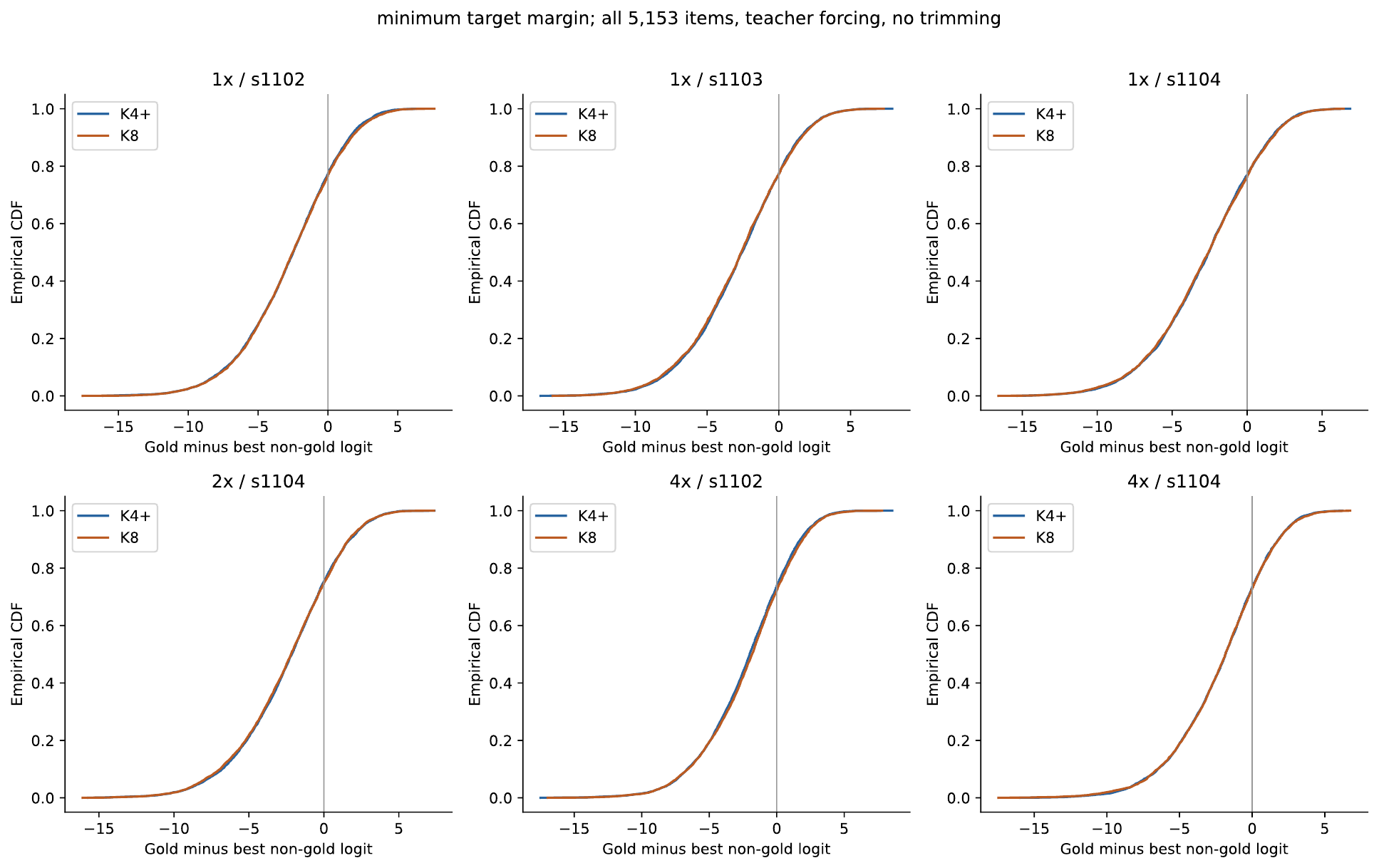}
\caption{Minimum teacher-forced target-token margin ECDFs across the same six pairs. Group-conditioned summaries and ties are retained in the numerical supplement; correctness-based separation alone is not independent mechanism evidence.}
\label{fig:marginmin}
\end{figure*}

\end{document}